%% file: main.tex
\documentclass[11pt,letterpaper]{article}

\usepackage[margin=1in]{geometry}
\usepackage{microtype}
\usepackage{amsmath,amsfonts}
\usepackage{algorithmic}
\usepackage{algorithm}
\usepackage{array}
\usepackage{textcomp}
\usepackage{url}
\usepackage{verbatim}
\usepackage{graphicx}
\usepackage{booktabs}
\usepackage{enumitem}
\usepackage{makecell}
\usepackage{colortbl}
\usepackage{multirow}
\usepackage{xcolor}
\definecolor{graybg}{gray}{0.95}
\usepackage{tabularx}
\usepackage{tikz}
\usepackage{subfigure}
\usepackage{pifont}
\usepackage[normalem]{ulem}
\usepackage{wrapfig}
\usepackage[colorlinks=true,linkcolor=blue,citecolor=blue,urlcolor=blue,bookmarks=true]{hyperref}

\newcommand{\mycirc}{\ding{108}}
\newcommand{\mydowntri}{\ding{116}}
\newcommand{\mysquare}{\ding{110}}
\newcommand{\myuptri}{\ding{115}}
\newcommand{\myx}{\ding{53}}
\newcommand{\mydiamond}{\ding{117}}
\newcommand{\myplus}{$\boldsymbol{+}$}

\input{math.tex}

\title{A Full-Pipeline Framework \\ for Evaluating Membership Inference Attacks in Machine Learning}

\author{
  Ding Chen\textsuperscript{2,1},
  Xinwen Cheng\textsuperscript{3,1},
  Xuyang Zhong\textsuperscript{2,1},
  Xinping Chen\textsuperscript{2},
  Xiaolin Huang\textsuperscript{3,4},
  Chen Liu\textsuperscript{2,4}\\[6pt]
  \textsuperscript{2}~City University of Hong Kong,\quad
  \textsuperscript{3}~Shanghai Jiao Tong University\\[6pt]
  {\small\texttt{ding.chen@my.cityu.edu.hk, xinwencheng@sjtu.edu.cn,}}\\
  {\small\texttt{xuyang.zhong@my.cityu.edu.hk, xinpichen2-c@my.cityu.edu.hk,}}\\
  {\small\texttt{xiaolinhuang@sjtu.edu.cn, chen.liu@cityu.edu.hk}}\\[6pt]
  \textsuperscript{1}~denotes equal contribution.\quad
  \textsuperscript{4}~denotes the correspondence author.
}
\date{}

\begin{document}

\maketitle

\begin{abstract}
While Membership Inference Attacks (MIAs) are the prevailing method for identifying training data, their application has expanded into privacy auditing and machine unlearning. Nevertheless, the field lacks a systematic framework for evaluating how different contexts affect MIA efficacy. Without such a characterization, practitioners risk deploying algorithms that perform well on benchmarks but become statistically irrelevant when faced with the nuances of specific, real-world datasets.
To bridge this gap and provide actionable insights, we introduce a comprehensive evaluation framework that systematically characterizes privacy risks across the entire machine learning pipeline, spanning data, architectures, algorithms, and post-training modules. Designed to inherently capture diverse operational contexts, our framework rigorously evaluates state-of-the-art MIAs across a broad spectrum of training configurations. To account for varying misclassification costs in real-world deployments, we employ three complementary metrics: Balanced Accuracy for symmetric costs, alongside TPR at low FPR (or TNR at low FNR) for asymmetric scenarios where false alarms or missed detections are strictly penalized. Furthermore, recognizing that existing MIAs assume divergent adversary capabilities, we formalize two standardized threat models and adapt these attacks into corresponding variants to ensure an equitable benchmark. Extensive empirical evaluations demonstrate that the efficacy of specific MIA methodologies is highly sensitive to the assumed threat models and chosen evaluation metrics. Ultimately, we distill these findings into actionable guidelines and provide a ready-to-use auditing toolkit, empowering practitioners to conduct better privacy assessments.
\end{abstract}


\input{Sections/intro}

\input{Sections/related_work}

\input{Sections/methodology}

\input{Sections/evaluation}

\input{Sections/conclusion}

\bibliographystyle{plain}
\bibliography{bib}

\appendix
\input{appendix/appendix.tex}

\end{document}

%% file: math.tex
\usepackage{amsmath,amsfonts,bm}

\def\eqref#1{equation~\ref{#1}}
\def\1{\bm{1}}

\def\vz{{\bm{z}}}

\DeclareMathAlphabet{\mathsfit}{\encodingdefault}{\sfdefault}{m}{sl}
\SetMathAlphabet{\mathsfit}{bold}{\encodingdefault}{\sfdefault}{bx}{n}

\def\gD{{\mathcal{D}}}

\def\gK{{\mathcal{K}}}

\def\gS{{\mathcal{S}}}

%% file: Sections/intro.tex
\section{Introduction}
The past decade has witnessed rapid advancements in machine learning. 
Overparameterized deep neural networks~\cite{vaswani2017attention, devlin-etal-2019-bert, achiam2023gpt} have pushed the boundaries of numerous application domains. However, unlike classical machine learning methods, whose solutions can often be expressed analytically, training deep neural networks involves solving ultra-high-dimensional nonconvex optimization problems, where the optima can only be approximated through numerical methods.
As a result, despite their impressive performance, these models are often regarded as black boxes with limited interpretability. In other words, deep neural networks may rely on feature representations vastly different from those used by humans, leading to various reliability concerns~\cite{goodfellow2014explaining, zhang2018visual, zhu2019deep}.

In this work, we focus on the issue of membership inference attacks (MIA)~\cite{shokri2017membership}, which infers whether or not a given data instance was included in the training set for a model.
Formally, consider a target machine learning model $f_\theta: \mathcal{X} \rightarrow \mathcal{Y}$ trained on a private dataset $\mathcal{D}_{\text{train}} = \{(x_i, y_i)\}_{i=1}^N$, where $\theta$ represents the model parameters. 
The goal of MIA is to determine whether a specific target sample $(x, y)$ was used to train the model $f_\theta$. 
Formally, this can be modeled as a binary hypothesis testing problem:
\begin{align*}
    \mathcal{H}_0 &: (x, y) \notin \mathcal{D}_{\text{train}} \quad (\text{Non-member}) \\
    \mathcal{H}_1 &: (x, y) \in \mathcal{D}_{\text{train}} \quad (\text{Member})
\end{align*}
The adversary $\mathcal{A}$ aims to construct an inference function that outputs a membership score $S_\text{{mem}} \in [0, 1]$, approximating the posterior probability $\Pr((x, y) \in \mathcal{D}_{\text{train}} \mid x, y, f_\theta)$.
Beyond its role as a threat, MIA has emerged as a \textit{de facto} standard for privacy auditing. For example, MIA is broadly used to evaluate the performance of unlearning algorithm~\cite{fan2024salun,huang2024unified}. For good MIA methods, a higher attack success rate directly quantifies the severity of data leakage, making MIA a critical metric for assessing the privacy-preserving efficacy of ML algorithms.

Despite the proliferation of MIA methodologies~\cite{shokri2017membership, salem2018ml, jayaraman2020revisiting, hui2021practical, song2021systematic}, a significant challenge persists for both adversaries and privacy auditors: identifying the optimal inference strategy under varying operational constraints. Current comparative studies suffer from two fundamental limitations. First, evaluations are often fragmented by conflated threat models. For example, classical approaches~\cite{song2021systematic} assume adversaries lack ground-truth data, whereas advanced methods\cite{carlini2022membership} leverage the ground-truth label to evaluate model's worst-case privacy leakage under their methods. Therefore, direct comparisons across these disparate settings lead to skewed risk assessments. 
Second, while recent benchmarks~\cite{niu2024survey,yeom2018privacy} offer valuable insights, their primary objective remains optimizing attack efficacy against static models within simplified pipelines. This fundamental divergence in goals leaves practitioners without actionable deployment strategies tailored to highly variable, real-world ML pipeline.

To resolve these issues, we introduce a full-pipeline framework to systematically evaluate the performance of MIA methods under various scenarios. First, we introduce two threat models: \textbf{Audit Mode} and \textbf{Attack Mode}, where the former has the full access to the ground-truth membership labels to optimize MIA settings. We design a standardized procedure to fairly compare existing MIAs under the same mode. Second, to accommodate the expanding scope of MIA applications, we equip our framework with tailored, asymmetric evaluation metrics, ensuring it flexibly fits a broad spectrum of real-world scenarios, ranging from strict privacy compliance to high-recall copyright screening, and their corresponding privacy requirements (e.g., TPR at low FPR or TNR at low FNR). Finally, building upon these settings, our framework decomposes the machine learning pipeline into discrete, configurable modules: data processing, model architectures, training algorithms, and post-training modifications. Crucially, our study dedicates an in-depth analysis to each module to isolate its unique impact on privacy leakage. For example, by adjusting the configurations within each module, we systematically dissect how data imperfections (e.g., mislabeling), architectural complexities, and defensive interventions (e.g., DP-SGD and machine unlearning) affect the MIA performance. Drawing on these empirical findings, we provide actionable takeaways for practitioners on how to select appropriate MIA methods under various situations.
The main contributions of this paper are summarized as follows:
\begin{itemize}[leftmargin=*]
    \item \textbf{Rigorous Evaluation Methodology:} We formalize two distinct threat models (\textit{Audit Mode} and \textit{Attack Mode}) based on the availability of the membership labels and establish a standardized protocol for strictly fair comparison of MIAs. Furthermore, we introduce tailored, asymmetric performance metrics (e.g., TPR at low FPR or TNR at low FNR) to accommodate diverse scenarios.

    \item \textbf{Full-Pipeline Evaluation Framework:} 
    We introduce a full-pipeline evaluation framework that integrates data processing, model training (including privacy-preserving techniques), and various post-training algorithms. This unified platform enables researchers to isolate confounding variables and systematically analyze the impact of individual pipeline components on overall privacy leakage.

    \item \textbf{Extensive Empirical Benchmark:} We conduct a large-scale benchmark of MIA methods across diverse datasets, architectures and training configurations. By moving beyond isolated scenario testing, this holistic evaluation provides empirical evidence on the comparative efficacy and practical limitations of state-of-the-art attack methodologies. Supported by extensive empirical results, we conclude practical guidelines on how to pick MIA methods under various situations.
    
\end{itemize}

%% file: Sections/related_work.tex
\section{Related Works} \label{sec:relatedworks}

\subsection{MIA Methods}
\begin{table}[ht]
    \centering
    \footnotesize
    \caption{Taxonomy of MIA methods based on the unified framework components. This table illustrates how distinct attacks instantiate the Representation Function ($v$), Auxiliary Knowledge ($\mathcal{K}$), and Classifier ($C$).}
    \label{tab:method_taxonomy}
    \renewcommand{\arraystretch}{1.2}
    \setlength{\tabcolsep}{3.5pt}
    \begin{tabular}{llll}
    \toprule
    \textbf{Method} & \textbf{Repr.~Function ($v$)} & \textbf{Aux.~Knowledge ($\mathcal{K}$)} & \textbf{Classifier ($C$)} \\
    \midrule
    \textbf{ShadowMIA} \cite{shokri2017membership} & Prediction Vector (Posteriors) & Shadow Models  & Parametric  \\
    \textbf{MLLeaks1} \cite{salem2018ml} & Prediction Vector (Top-k) & Shadow Models & Parametric \\
    \textbf{MLLeaks3} \cite{salem2018ml} & Scalar Metric  & Reference Data & Static Threshold \\
    \textbf{Merlin} \cite{jayaraman2020revisiting} & Loss Landscape  & Shadow Models  & Static Threshold \\
    \textbf{Metric MIA} \cite{song2021systematic} & Labelwise Scalar Metric & Shadow Models  & Labelwise Static Threshold \\
    \textbf{LiRA} \cite{carlini2022membership} & Likelihood Ratio  & Shadow Models  & Static Threshold \\
    \textbf{Quantile MIA} \cite{bertran2024scalable} & Scalar Metric (Loss) &Reference Data & Adaptive Threshold \\
    \textbf{RMIA} \cite{zarifzadeh2024low} & Pairwise Likelihood Ratio & Shadow Models & Adaptive Threshold \\
    \textbf{Blind-MI} \cite{hui2021practical} & Differential Shift (Set-based) & Reference Data & Geometric Distance \\
    \bottomrule
    \end{tabular}
\end{table}

We propose the following unified framework to systematically analyze different MIA methods that predict the membership $S_{\text{mem}}$ of $x$ for model $f_\theta$:
\begin{equation}
    \label{eq:unified_framework}
    S_{\text{mem}} = C\big(v(x, f_\theta; \mathcal{K})\big)
\end{equation}
where $v(\cdot)$ denotes the representation function responsible for signal extraction, $\mathcal{K}$ represents the auxiliary knowledge available to the adversary and $C(\cdot)$ acts as the classifier to discriminate between members and non-members. Within this framework, existing MIA methodologies can be systematically characterized by their distinct instantiations of these components. We summarize the components of different representative MIA methods in Table~\ref{tab:method_taxonomy}.

\noindent \textbf{Representation Function.}
The function $v$ is responsible for extracting the \textit{MIA score} $\mathbf{z}$ from the interaction between the target sample, the model and the auxiliary knowledge. 
It captures the behavioral traces of the sample. Generally speaking, three kinds of representation function are popular among MIA methods:

\begin{itemize}[leftmargin=*]
    \item \textit{Logit / Posterior Vectors.} 
    Seminal works, such as ShadowMIA \cite{shokri2017membership} and MLLeak1\cite{salem2018ml}, directly utilize the full confidence vector (softmax posteriors) or the raw logits as the MIA score $\mathbf{z}$. 

    \item \textit{Scalar Metrics.} 
    Many methods compress the model output into scalar metrics to reduce the feature dimensionality.
    Approaches like Quantile MIA \cite{bertran2024scalable}, MLLeak3 \cite{salem2018ml} and Metric MIA \cite{hui2021practical} construct the MIA score $\vz$ by various statistical aggregates, including cross-entropy, modified entropy, maximum posterior probability (confidence) and standard deviation.

    \item \textit{Calibrated Metrics.}
    The calibrated metrics, such as the calibrated likelihood ratio~\cite{carlini2022membership}, are proposed recently to mitigate the influence of ``sample hardness'': an easy non-member might exhibit a lower loss than a hard member.
    Compared with scalar metrics, calibrated metrics are ``calibrated'' by a reference model parameterized by $\theta_{\text{ref}}$, which are usually derived from shadow models or population statistics.
    Formally, calibrated likelihood ratio in log-space is $v(x, f_\theta, \gK) = \log \frac{\Pr(x|f_\theta)}{\Pr(x|f_{\theta_{\text{ref}})}}$.

    Moreover, \cite{zarifzadeh2024low} extends it to pairwise likelihood ratio tests.

    \item \textit{Loss Landscape Measure.} Merlin \cite{jayaraman2020revisiting} assumes that member data are more likely to be near local minima than non-member records, which suggests that for members, loss is more likely to increase at perturbed points near the original, whereas it
    is equally likely to increase or decrease for non-members. Hence, it adds small perturbation to the
    input and uses the output differences as score to distinguish the member and non-member data.
\end{itemize}

\noindent \textbf{Auxiliary Knowledge.} Auxiliary knowledge $\gK$ defines the adversary's capability to access external resources, which usually comprises a dataset $\gD_{\text{aux}}$ drawn from the same underlying distribution as the private training data.
In literature, $\mathcal{K}$ is utilized in following two primary ways:

\begin{itemize}[leftmargin=*]
    \item \textit{Shadow Modeling.} Many attack methods such as Shadow \cite{shokri2017membership}, MLLeaks \cite{salem2018ml}, Metirc-MIA \cite{song2021systematic}, LiRA \cite{carlini2022membership} and RMIA \cite{zarifzadeh2024low} leverage $\mathcal{K}$ to train \textit{shadow models}, that mimic the target model's behavior. 
    By partitioning $\mathcal{D}_{\text{aux}}$ into shadow ``member'' and ``non-member'' sets, the adversary utilizes these models in two distinct ways: 
    (1) to generate labeled $(\mathbf{z}, \gS_{\text{mem}})$ pairs for training the classifier in a supervised manner; 
    (2) to statistically estimate the non-member reference distribution, which serves as the critical denominator for calculating calibrated likelihood ratio scores.

   \item \textit{Reference Data.} Some attack methods such as Quantile MIA\cite{bertran2024scalable}, MLLeak3 \cite{salem2018ml} and BlindMI \cite{hui2021practical} directly generate the non-member data  $\mathcal{D}_{\text{aux}}$ acts as non-member samples to estimate the baseline behavior of the model on unseen data rather than use it for shadow modeling training.
\end{itemize}

\noindent \textbf{Classifier.} The function $C$ mapping $\vz$ to either \text{member} or \text{non-member} serves as the Inference Classifier. 
The classifier $C$ interprets features $\vz$ to produce the final binary decision. 
Existing literature instantiates $C$ using varying degrees of complexity, ranging from simple static thresholds to sophisticated learning-based models:

\begin{itemize}[leftmargin=*]
    \item \textit{Parametric Classifiers.} 
    When the representation $\mathbf{z}$ is high-dimensional (e.g., logit or posterior vectors), $C$ is typically instantiated as a parametric binary classifier like a SVM or a multi-layer perceptron. 
    As demonstrated in ShadowMIA and MLLeaks \cite{shokri2017membership}, the adversary utilizes the shadow modeling technique to generate a labeled dataset of member and non-member features, and then train the classifier $C$ in a supervised manner to learn the non-linear decision boundary that separates members from non-members.

    \item \textit{Global Static Thresholding.} 
    For MIA using scalar metrics like Metric MIA \cite{song2021systematic}, MLLeak3 \cite{salem2018ml}, Merlin \cite{jayaraman2020revisiting} and LiRA \cite{carlini2022membership}, the simplest form of classification is a global thresholding function $\mathbb{I}(\mathbf{z} < \tau)$, which can also be considered as a 1-dimensional linear classifier.
    The threshold $\tau$ is typically selected based on the reference distribution estimation.
    \item \textit{Sample-specific Adaptive Thresholding.} Recognizing that ``sample hardness'' varies, recent methods employ adaptive thresholding which dynamically generates a threshold $\tau(x)$ given a target $x$.
    Notably, Quantile MIA \cite{bertran2024scalable} trains a quantile regression model on $\mathcal{D}_{\text{aux}}$ to predict the distribution of non-member signals conditional on the input sample. 
    \item \textit{Geometric Approaches.} 
    Unlike the aforementioned methods that classify samples independently, Blind-MI \cite{hui2021practical} predicts sample membership in batch from a geometric perspective. 
    It operates on the insight that the inclusion of a non-member into a dataset shifts the dataset's centroid in the high-dimensional feature space towards the non-member distribution. 
    Here, $C$ utilizes $\mathcal{D}_{\text{aux}}$ as a reference anchor and leverages differential comparison (e.g., distance metrics) to statistically distinguish the member subset from the non-member subset within the target batch.
\end{itemize}

\subsection{MIA Surveys}

As the landscape of membership inference attacks (MIAs) rapidly expands, several benchmarking efforts have sought to systematize these methods to understand their real-world threats. Early comprehensive studies, such as ML-DOCTOR~\cite{liu2022ml}, conducted holistic risk assessments across multiple privacy attacks, including MIA, model inversion, and model stealing. These works successfully demonstrated that macroscopic factors, such as dataset complexity and model overfitting, significantly dictate attack performance.

A persistent challenge in MIA literature, however, is the prevalence of \textit{Conflicting Comparison Results (CCR)} stemming from inconsistent experimental setups and evaluation metrics. To address this, frameworks like MIBench~\cite{niu2025comparing} introduced standardized evaluation scenarios, though their scope remained strictly focused on data-level factors. Concurrently, other studies~\cite{wang2025membership} have shifted focus from overarching performance metrics (e.g., AUC) toward the fine-grained reliability and disparity of attacks on, revealing that MIAs with similar aggregate success rates often exploit entirely different subsets of vulnerable samples. Additionally, while there is growing interest in benchmarking MIAs against Large Language Models (LLMs)~\cite{mireshghallah2022quantifying, shi2023detecting}, these evaluations fundamentally rely on sequence-level memorization metrics (e.g., perplexity or exact string matching). These are inherently incompatible with the instance-level, probability-based metrics used in standard classification tasks, preventing a unified comparison.

This work focuses on classification tasks for a unified and consistent comparison. Compared with existing benchmarks which conflate threat models, optimize against static models or restrict focus to a specific stage of machine learning process, this work bridges these gaps by moving beyond isolated scenario testing to introduce a comprehensive, full-pipeline evaluation framework.

%% file: Sections/methodology.tex
\section{Unified Evaluation Protocol} \label{sec:protocol}

In this section, we establish a unified protocol to ensure equitable comparisons across Membership Inference Attacks (MIAs). We first formalize two distinct threat models to standardize adversarial capabilities. Subsequently, we detail a comprehensive suite of metrics designed to rigorously assess attack efficacy under diverse operational constraints.

\subsection{Threat Model}
\label{subsec:threat_model_discussion}

MIA considers an adversary with black-box query access to the target model, who also possesses an auxiliary dataset $\mathcal{D}_{\text{aux}}$ drawn from the same underlying distribution as the private training set $\mathcal{D}_{\text{train}}$. Within this framework, we explicitly distinguish between two operational scenarios based on the adversary's privileges: the \textbf{audit mode} and the \textbf{attack mode}. 
In the audit mode, a privacy auditor aims to quantify the theoretical worst-case leakage risks from the model under specific MIA method. Therefore, the auditor is granted access to ground-truth membership labels, i.e., whether or not the samples are in $\gD_\text{train}$. This access allows the auditor to derive optimal hyper parameters. 
By contrast, in the attack mode, a malicious attacker seeks to compromise privacy in a realistic setting without access to ground-truth labels. Therefore, it solely relies on $\mathcal{D}_{\text{aux}}$ to estimate attack hyper parameters, through ways like training shadow models on $\mathcal{D}_{\text{aux}}$.

Distinguishing the audit mode and the attack mode is necessary for fair comparison among different MIA methods. Pioneering works such as ShadowMIA\cite{shokri2017membership}, MetricMIA\cite{song2021systematic}, BlindMI\cite{hui2021practical}, are developed under the attack mode. However, Recent methods such as LiRA\cite{carlini2022membership}, Quantile MIA\cite{bertran2024scalable}, RMIA\cite{zarifzadeh2024low} are all based on the audit mode, causing discrepancy in threat models. 
To facilitate an equitable comparison across MIA methods, we construct corresponding variants for each algorithm, enabling their evaluation under both audit and attack modes.

\noindent \textbf{From Attack to Audit.}
Methods such as MetricMIA~\cite{hui2021practical} and Merlin~\cite{jayaraman2020revisiting} were natively designed for the attack mode, relying on heuristic metrics without ground-truth supervision. 
To evaluate their theoretical limits in the audit mode, we formulate their \textbf{Oracle Variants}. 
Leveraging the auditor's access to ground-truth membership labels, we replace the static or heuristic hyper parameters of these methods with empirically optimal ones. 
This adaptation allows us to quantify the worst-case leakage risk from the model by these methods.

\noindent \textbf{From Audit to Attack.}
Conversely, advanced baselines like LiRA~\cite{carlini2022membership} and RMIA~\cite{zarifzadeh2024low} function natively in the audit mode, often presuming access to ground-truth labels to calibrate scores or utilizing extensive computational resources impractical for real-time attacks. 
To assess their feasibility in the attack mode, we develop their \textbf{Shadow Variants}. 
In the attack mode, the attacker has no access to membership labels, which makes us pick hyper parameters solely from the auxiliary dataset $\mathcal{D}_{\text{aux}}$ via shadow models.
Specifically, we train several shadow models on $\mathcal{D}_{\text{aux}}$ and pick the optimal hyper parameters for MIA using $\mathcal{D}_{\text{aux}}$ and these models.

For MIA methods that do not require hyper parameter tuning, the attack and audit modes are functionally identical. For tunable MIA methods, the discrepancy in performance between these modes quantifies the method's stability and its sensitivity to hyper parameter optimization.

\subsection{Metrics}

To comprehensively assess the privacy risks quantified under the \textit{audit} and \textit{attack} modes defined above, we employ a multi-dimensional evaluation strategy.

We first adopt \textit{balanced accuracy} to provide a holistic view of membership distinguishability. Specifically, we evaluate the attack accuracy on a balanced dataset sampled with an equal number of members and non-members.
However, aggregate metrics like balanced accuracy often mask the severity of leakage for vulnerable outliers~\cite{carlini2022membership}. Therefore, we prioritize two regime-specific metrics that align with distinct realistic requirements in practice:

\noindent \textbf{High-Confidence Leakage (TPR at Low FPR).}
This metric is the primary standard for the audit mode. In high-stakes domains (e.g., medical records or financial data), the privacy risk is defined by the adversary's ability to identify \textit{any} specific individual instance with high certainty. 
Following the existing rigorous audit standards~\cite{carlini2022membership}, we measure the true positive rate (TPR) at a strictly limited false positive rate (FPR) threshold (e.g., only 0.1\% non-members are misclassified). 
A non-trivial TPR at such low FPR implies that the method can confidently isolate a subset of $\mathcal{D}_{train}$ without incurring significant false accusations. 

\noindent \textbf{High-Recall Screening (TNR at Low FNR).}
Complementing the high-confidence metric, we introduce true negative rate (TNR) at low false negative rate (FNR) to evaluate scenarios prioritizing coverage, such as copyright enforcement or dataset cleansing\cite{shi2023detecting}. 
Consider a copyright owner seeking to identify all potentially infringed samples contained within a suspect model, or the model developer attempts to sanitize the training corpus by removing unauthorized or sensitive samples, a missed detection (false negative) is more costly than a false alarm. 
In this context, we evaluate the method's capability to filter out non-members (High TNR) conditioned on a low FNR (e.g., 99.9\% of members are recalled). 
This metric assesses the utility of MIA as a screening tool, balancing the rigorous retrieval of member samples with the workload required for subsequent verification.

In summary, the selection of these three metrics is driven by the varying misclassification costs in different real-world applications. Specifically, TPR at low FPR and TNR at low FNR are adopted for scenarios with asymmetric costs, where the penalties for false alarms (false positives) or missed detections (false negatives) are disproportionately high, respectively. In contrast, balanced accuracy serves as a standard measure for scenarios with symmetric costs, where false positives and false negatives carry roughly equivalent consequences. By integrating balanced accuracy with the two regime-specific metrics, our framework provides a comprehensive assessment of privacy risks across varying operational requirements.

\subsection{Full-Pipeline Benchmark Framework}
\label{sec:pipeline}

\begin{wrapfigure}{r}{0.48\textwidth}
    \centering
    \includegraphics[width=\linewidth]{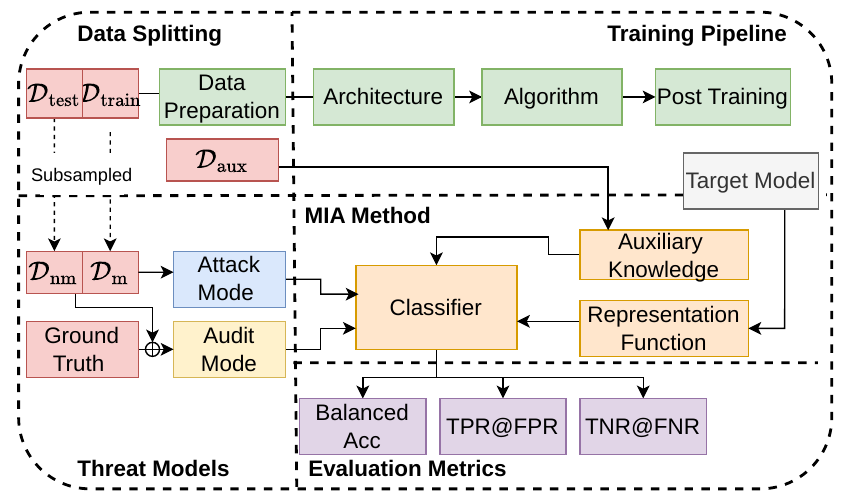}
    \caption{The full-pipeline framework. This pipeline consists of four primary stages. For a given dataset, we evaluate how
    MIA performance across data preparation, model architecture, training algorithms, and post-training scenarios.}
    \label{fig:framework}
\end{wrapfigure}
To rigorously benchmark Membership Inference Attacks (MIAs), Figure \ref{fig:framework} formulates an framework where the analytical pipeline flows continuously from data configuration to final assessment across intrinsically linked modules. The process initiates with the \textbf{Data Splitting} module, which partitions the source dataset $\mathcal{D}_{\text{source}}$ into two strictly disjoint and identically distributed subsets: the target data and the auxiliary data. The target data comprises $\mathcal{D}_{\text{train}}$ (exclusively used to train the target model $f_\theta$) and $\mathcal{D}_{\text{test}}$, from which we further subsample the member ($\mathcal{D}_{m} \subseteq \mathcal{D}_{\text{train}}$) and non-member ($\mathcal{D}_{nm} \subseteq \mathcal{D}_{\text{test}}$) datasets. The accessibility of these ground-truth membership is governed by the \textbf{Threat Models} module, which defines the boundaries of adversarial capabilities. Specifically, the auditing mode is granted oracle access to these labels, whereas the attack mode operates entirely without such knowledge.

Building upon modules, we have the \textbf{Training Pipeline} module to train models. The pipeline consists of three stages, covering the entire lifecycle from data preparation to post-training: (i) architecture configuration, (ii) training process, and (iii) post-training. 

Operating within these defined privilege constraints, the \textbf{MIA Methods} module is deployed to execute the attack by leveraging the remaining auxiliary dataset ($\mathcal{D}_{\text{aux}} = \mathcal{D}_{\text{aux}}^{\text{train}} \cup \mathcal{D}_{\text{aux}}^{\text{test}}$) as the adversary's background knowledge. Specifically, in scenarios where shadow models are required, such as in the attack mode, we train $5$ shadow models. The training data (member) and the test data (non-member) are sampled from $\gD_{\text{aux}}^{\text{train}}$ and $\gD_{\text{aux}}^{\text{test}}$, respectively, using a sampling rate $r = 0.8$ to ensure diversity. Ultimately, the predictive outputs generated by these formulated MIA methods are ingested directly into the \textbf{Evaluation Metrics} module, which systematically quantifies the attack efficacy against the defined ground truth, thereby completing the evaluation lifecycle. The pseudo-code is demonstrated in Algorithm~\ref{alg:mia}.

\begin{algorithm}[h!]
\caption{Pipeline of MIA with \colorbox{red!35}{Attack} / \colorbox{green!35}{Audit} Mode.}
\label{alg:mia}
\begin{algorithmic}[1]
    \REQUIRE MIA method $\mathcal{A}$; Datasets $\gD_{\text{train}}$, $\gD_{\text{test}}$, $\gD_{\text{aux}}=\mathcal{D}_{\text{aux}}^{\text{train}} \cup \mathcal{D}_{\text{aux}}^{\text{test}}$; Number of shadow models $N$;  Sampling rate for sub-shadow dataset $r=0.8$.\\ 
    \textit{\textbf{\# Select model architecture and model training}}
    \STATE Train target model $f_\theta$ on $\gD_{\text{train}}$ \\
    \textit{\textbf{\# Train shadow models if applicable}} 
    \FOR{$i$ in $1,~...,~N$}
        \STATE $\gD_{\text{sm}_{i}} \leftarrow$ Sample $\gD_{\text{aux}}^{\text{train}}$ with sampling rate $r$
        \STATE $\gD_{\text{sn}_{i}} \leftarrow$ Sample $\gD_{\text{aux}}^{\text{test}}$ with sampling rate $r$
        \STATE Train shadow model $f_{\text{sm}_i}$ on $\gD_{\text{sm}_{i}}$
    \ENDFOR\\
    \textit{\textbf{\# Hyperparameter tuning under different threat Model}}

    \STATE Set hyper parameters of $\mathcal{A}$ by \colorbox{red!35}{$\{(f_{\text{sm}_i}, \gD_{\text{sm}_i}, \gD_{\text{sn}_i})\}_{i = 1}^N$} or \colorbox{green!35}{$(f_\theta, \gD_{\text{train}}, \gD_{\text{test}})$} \\
    \textit{\textbf{\# Evaluation}}
    \STATE Get the prediction results of $\gD_{\text{train}}$, $\gD_{\text{test}}$  using $\mathcal{A}$.
    \ENSURE Prediction results (member/non-member)\\
\end{algorithmic}
\end{algorithm}

In the subsequent section, we systematically investigate the efficacy of the formulated MIA methods in alignment with the data preparation phase and each phase of this training pipeline. Establishing a ResNet-18 architecture trained on the CIFAR-100 dataset as our standard baseline, we conduct comprehensive empirical evaluations. By independently altering individual configurations within the data preparation and training pipeline stages while strictly holding all other settings fixed at the baseline, we thoroughly analyze the performance dynamics of various MIAs.

%% file: Sections/evaluation.tex
\section{Evaluation and Observation}
\label{sec:exp_results}

In this section, we follow the execution protocol in Algorithm~\ref{alg:mia} to study how the modules discussed in Section~\ref{sec:pipeline} affect the performance of MIA methods. Unless otherwise stated in specific ablation studies, we use ResNet-18 trained on CIFAR100 as the reference pipeline for comparison.

\subsection{Impact of Threat Models}
The hyper parameters of MIA methods are mainly decision thresholds.
Theoretically, the audit mode establishes the best performance by utilizing ground-truth supervision to derive the optimal thresholds. In contrast, the attack mode reflects realistic risks constrained by the adversary's estimation errors, relying solely on shadow models to approximate these thresholds. We utilize the balanced accuracy gap between these two modes to quantify the performance stability of an MIA method in response to threshold miscalibration. A narrower gap demonstrates superior performance stability and practical reliability in real-world settings.

\begin{table}[ht]
\centering
\caption{Audit-Attack Gap on Different Scenarios}
\label{tab:gap_sensitivity_with_lines}
\scriptsize
\resizebox{\columnwidth}{!}{
\setlength{\tabcolsep}{2pt}
\begin{tabular}{l@{\hspace{1.5mm}}c@{\hspace{1.5mm}}c@{\hspace{1.5mm}}c@{\hspace{1.5mm}}c@{\hspace{1.5mm}}c@{\hspace{1.5mm}}c@{\hspace{1.5mm}}c@{\hspace{1.5mm}}c}
\Xhline{1.5pt}
\multirow{2}{*}{Scenario} & \multirow{2}{*}{Set} & \multirow{2}{*}{LiRA} & \multirow{2}{*}{RMIA} & \multicolumn{3}{c}{Metric MIA} & \multirow{2}{*}{Quantile MIA} & \multirow{2}{*}{Merlin} \\
\cmidrule(lr){5-7}
 &  &  &  & Entropy & \makecell{Modified\\Entropy} & Confidence &  &  \\
\Xhline{1pt}
Baseline & CIFAR-100 & \makecell{74.15 \\ 67.45} & \makecell{75.26 \\ 74.25}          & \makecell{80.65 \\ 78.49} & \makecell{81.97 \\ 80.25} & \makecell{81.90 \\ 80.30} & \makecell{80.53 \\ 80.53}          & \makecell{56.51 \\ 56.36} \\
\midrule
Architecture & 4-CNN     & \makecell{63.66 \\ 55.58} & \makecell{63.44 \\ 63.31}          & \makecell{59.59 \\ 53.18} & \makecell{65.22 \\ 60.34} & \makecell{65.05 \\ 60.90} & \makecell{62.27 \\ 62.11}          & \makecell{54.37 \\ 50.11} \\
     & Swin-T    & \makecell{64.07 \\ 58.25} & \makecell{62.43 \\ 57.16} & \makecell{66.91 \\ 63.94}          & \makecell{67.93 \\ 65.01} & \makecell{67.84 \\ 64.78} & \makecell{65.16 \\ 65.24}         & \makecell{56.52 \\ 52.40} \\
\midrule
Dataset & CIFAR-10  & \makecell{56.19 \\ 52.76} & \makecell{56.53 \\ 53.61}          & \makecell{65.68 \\ 59.87} & \makecell{66.17 \\ 60.55} & \makecell{66.17 \\ 60.68} & \makecell{61.19 \\ 60.98}          & \makecell{56.96 \\ 56.45} \\
     & ImageNet100 & \makecell{60.34 \\ 59.66} & \makecell{60.56 \\ 60.30}          & \makecell{61.64 \\ 54.12} & \makecell{63.04 \\ 56.14} & \makecell{62.90 \\ 55.98} & \makecell{57.26 \\ 51.05} & \makecell{52.54 \\ 52.02} \\
\midrule
Algorithm  & DP-SGD        & \makecell{50.03 \\ 54.17} & \makecell{64.74 \\ 63.57}          & \makecell{64.32 \\ 61.04} & \makecell{68.11 \\ 64.58} & \makecell{68.07 \\ 64.60} & \makecell{66.52 \\ 65.93}  & \makecell{56.18 \\ 58.69} \\
\Xhline{1.5pt}
\end{tabular}
}
\begin{flushleft}
\scriptsize \textit{Note:} Upper values denote MIA methods' accuracy under the audit mode; lower values indicate the accuracy under the attack mode.
\end{flushleft}
\end{table}

Table \ref{tab:gap_sensitivity_with_lines} indicates that many methods suffer from a large audit-attack gap, demonstrating their high sensitivity to empirical distribution discrepancy between $\gD_{\text{train}}$ and $\gD_{\text{aux}}$. Specifically, across different target model architectures and datasets, we find that Metric MIAs, despite strong performance in the auditing mode, exhibit substantial performance degradation in the attack mode. By contrast, Quantile MIA turns out to be the most reliable choice, which maintains a minimal audit-attack gap across diverse operational conditions.

\subsection{Impact of Data}
In this subsection, we investigate the impact of distinct data characteristics on MIA susceptibility. Following the protocol defined in Algorithm~\ref{alg:mia}, we evaluate the influence of data on various datasets, including CIFAR10 \cite{krizhevsky2009learning}, CIFAR100\cite{krizhevsky2009learning} (with fine-grained or superclass labels), and ImageNet100. We use WideResNet-28 for ImageNet100\cite{ILSVRC15} and ResNet-18 for other datasets. Table~\ref{tab:data_complexity_comparison} summarizes the performance of various MIA methods across different datasets. 

\noindent \textbf{Impact of Dataset Complexity.}
Compared with the fine-grained CIFAR100, the superclass CIFAR100 poses greater challenges for MIA methods. This is because the model for the superclass CIFAR100 has lower-dimensional outputs, providing less information for MIAs. Additionally, when evaluating performance across different datasets, we observe a strong correlation between MIA vulnerability and generalization gaps, which are highlighted in Table \ref{tab:performance_summary} of Appendix \ref{subsec:app_mia_scratch}.
Based on the results in Table~\ref{tab:data_complexity_comparison}, we can see Metric MIA consistently has better performance than other MIAs in all metrics, including balanced accuracy, TPR at low FPR and TNR at low FNR.

\input{experiments/data_exp}

\noindent \textbf{Impact of Mislabeled Data.}

\begin{figure}[ht]
    \centering
    \includegraphics[width=0.45\textwidth]{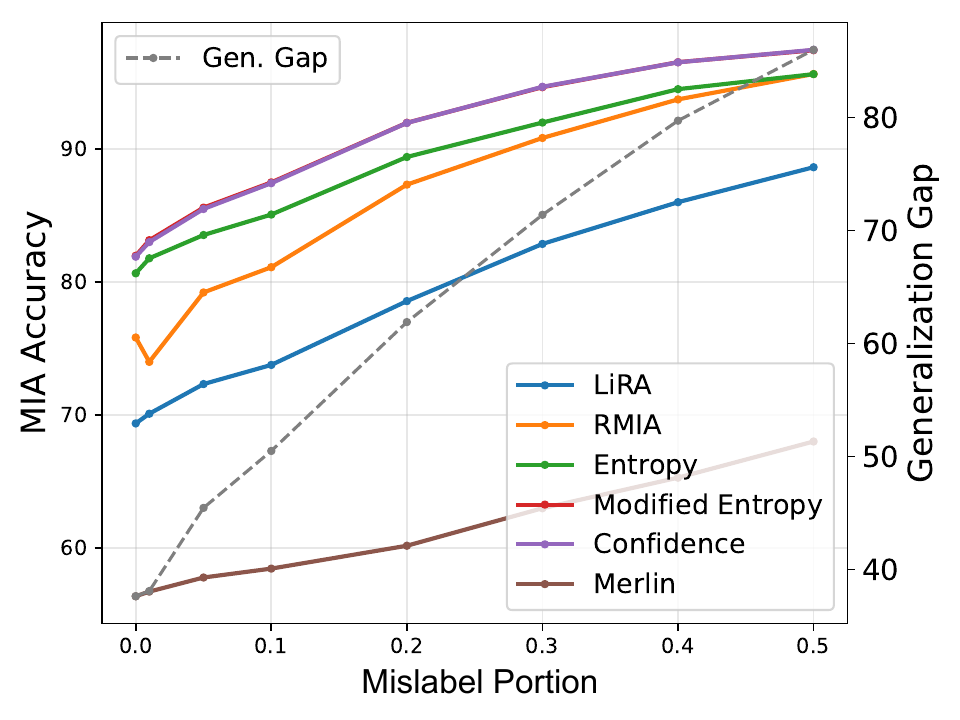}
    \vspace{-0.5em}
    \caption{\small The accuracy of MIA methods on CIFAR100 under Audit Mode, indicated by the left y-axis, with different mislabel portions: $0.01$, $0.05$, $0.1$, $0.2$, $0.3$, $0.4$, and $0.5$. Gray dashed lines indicate the generalization gap in each mislabeled portion, indicated by the right y-axis. The architecture is ResNet-18. We consider LiRA, RMIA, three variants of Metric MIA, and Merlin in the auditing mode. Due to prohibitive computational complexity, Quantile MIA is excluded from the evaluation.}
    \label{fig:mia_curve_mis}
\end{figure}

Data quality, particularly label correctness, plays a pivotal role in privacy defense. Theoretically, deep networks tend to ``memorize'' mislabeled examples (as outliers) rather than learning generalizable patterns from them \cite{zhang2017understanding}. 
To quantify this, we deliberately introduce symmetric label noise into the CIFAR100 baseline. We randomly corrupt 10\% of the training labels, forcing the model to overfit these confusing samples.
As shown in Table~\ref{tab:data_complexity_comparison}, the models trained with noisy labels are more susceptible to MIAs.
Furthermore, Figure~\ref{fig:mia_curve_mis} demonstrates that higher portion of mislabeled data leads to higher model vulnerability against MIAs.

\subsection{Impact of Architectures}

The architecture configuration identifies the model expressive power for both target and shadow models. We consider architectures of different scales and topologies, especially the realistic black-box scenarios where the shadow models have different architectures from the target model.

We examine a diverse range of neural network architectures varying in capacity and structure. Our selections include some representative topologies and depth, including a 4-layer CNN \cite{krizhevsky2012imagenet}, VGG-11 \cite{simonyan2015very}, ResNet-18/34/50 \cite{he2016deep}, and WideResNet-28-2 \cite{zagoruyko2016wide} and Swin Transformer-Tiny \cite{liu2021swin}.
 The results in Table \ref{tab:arch_comparison} show different vulnerability of these architecture configurations. However, Metric MIA consistently outperforms other MIA methods in all three metrics we consider, followed by Quantile MIA.

\input{experiments/architecture}

\subsection{Impact of Training Algorithms}
\label{subsec:impact_algo}

The choice of the training algorithm and the optimization strategy dictates how a model encodes information from the training data. 
In this context, we study the temporal dynamics of membership leakage during training and how privacy-preserving algorithms mitigate the MIA vulnerability.

\begin{figure}[htbp]
    \centering    \includegraphics[width=0.45\textwidth]{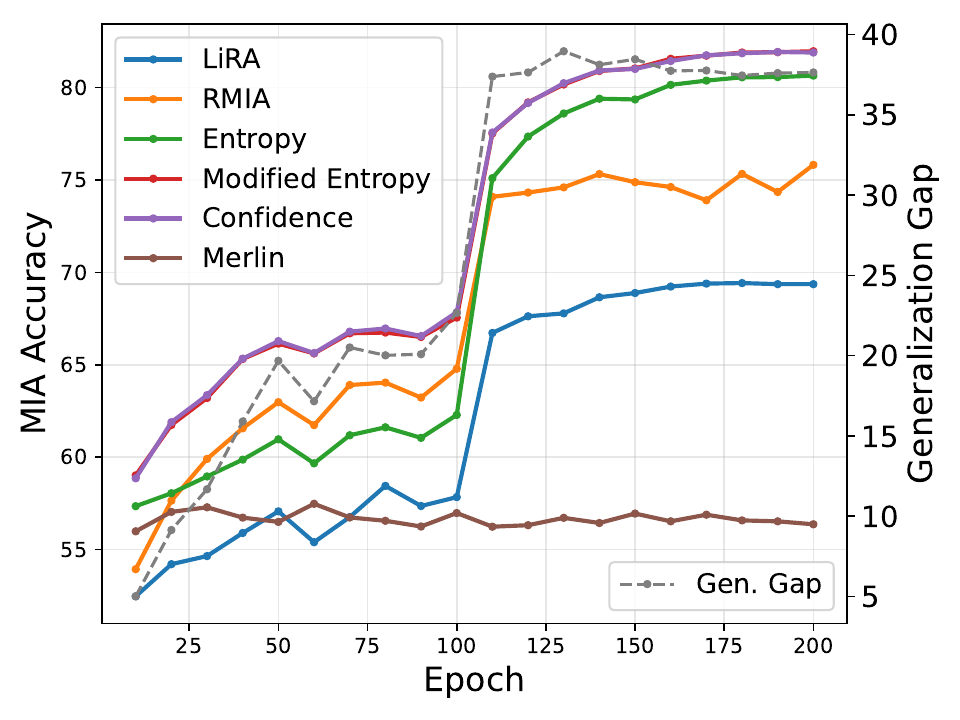}
    \vspace{-0.5em}
    \caption{\small MIA accuracy vs. training epoch. The evaluation is conducted every 10 epochs. Gray dashed lines indicate the generalization gap. LiRA, RMIA, three variants of Metric MIA, and Merlin in auditing mode are tested. Due to prohibitive computational complexity, Quantile MIA is not included in the evaluation. Note that Merlin exhibits trivial performance in both settings. }
    \label{fig:train_alg_mia_ep}
\end{figure}

\noindent \textbf{Training Progress.} We track performances of various MIAs throughout training process and report their performances at early, intermediate, and final stages. As illustrated in Figure~\ref{fig:train_alg_mia_ep}, performances of all the MIAs improve steadily as training progresses, suggesting that the  membership signals are clearer as model increasingly encodes information from the training data.
That is to say, MIA vulnerability gradually increases as training progresses. 
Among different MIA methods, Metric MIA using modified entropy or confidence consistently outperforms other MIAs during the whole training process.
In addition, The success rate of MIA methods has strong correlation with the generalization gap of the model, indicating overfitting models are more vulnerable against MIAs.

\noindent \textbf{Privacy-Enhancing Algorithms.} To investigate how much the privacy protection training algorithm mitigates privacy risk, we evaluate MIA performances in DP-SGD~\cite{abadi2016deep} trained model in Table~\ref{tab:train_alg}. We could see that almost all the MIAs exhibit worse performances in a DP-SGD trained model, validating the DP mechanism to protect the data privacy. 
Specifically, some competitive MIA methods under vanilla training like LiRA and ML-Leaks collapse to near-random guess under DP-SGD, indicating their poor adaptivity to DP noise.
Nevertheless, Metric MIA still stands out and achieves the best performance.

\subsection{Impact of Post Training} \label{subsec:post-train}

\begin{wrapfigure}{r}{0.45\textwidth}
    \centering
    \includegraphics[width=\linewidth]{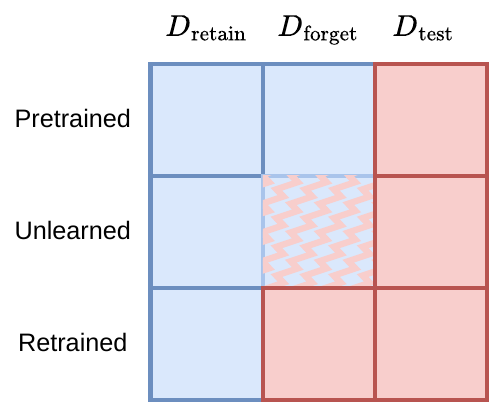}
    \caption{Membership across different models and datasets in unlearning: blue denotes members while red denotes non-members. The membership of $\mathcal{D}_{\text{forget}}$ is ambiguous and depends upon unlearning algorithms. Ideally, a perfect unlearning algorithm should produce a model indistinguishable from one that was retrained, effectively making $\mathcal{D}_{\text{forget}}$ non-members.}
    \label{fig:mu}
\end{wrapfigure}

Our pipeline considers the post training stage and specifically two distinct scenarios: (1) finetuning which aims to fit additional training data; (2) machine unlearning (MU) which aims to erase the influence of some existing training data. 

\noindent \textbf{Finetuning.} For finetuning tasks, we use the dataset $\gD^{\text{train}}_{\text{ft}}$, which has no overlap with the pretraining set $\gD_{\text{train}}$, to finetune the model with the same optimization settings as in the final phase of pretraining. To evaluate various MIAs, we define members as samples in $\gD^{\text{train}}_{\text{ft}}$ and non-members as samples from a holdout set $\gD^{\text{test}}_{\text{ft}}$, which follows the same distribution as $\gD^{\text{train}}_{\text{ft}}$.
In the experiments, we finetune an ImageNet-pretrained ResNet50 model on CIFAR100 and CIFAR10, respectively.
To investigate how different training regimes affect a model's susceptibility to MIAs, we compare models finetuned models with models trained from scratch on the same dataset. 

As shown in Table~\ref{tab:train_alg}, it is consistently more challenging for MIAs to targert a finetuned model compared with one trained from scratch. 
We attribute this to the smaller learning rates and gradient magnitudes during finetuning, which slow the incorporation of training-specific information and consequently produce weaker membership signals. 
Moreover, Quantile MIA achieves the best overall performance when targeting a fine-tuned model, highlighting its advantage in detecting weaker training signals. By contrast, most other MIA methods generate trivial performance near random guesses.

\input{experiments/train_alg}

\input{experiments/unlearn_result_reduced}

%% file: experiments/data_exp.tex
\begin{table*}[h!tbp]
\centering
\tabcolsep=0.18em 
\renewcommand{\arraystretch}{1.25} 
\tiny 
\caption{\small Performance of different MIA methods under different datasets. } 

\label{tab:data_complexity_comparison}
\newcommand{\resTwo}[3]{\makecell[c]{#1 \\ \color{gray}\tiny #2~/~#3}}

\newcommand{\resOne}[1]{\makecell[c]{#1 \\ \vphantom{\tiny 0.00~/~0.00} }} 

\resizebox{\textwidth}{!}{%
\begin{tabular}{cccccccccccccccc}
\Xhline{2\arrayrulewidth}

\multirow{3}{*}{\textbf{Data}}& \multicolumn{14}{c}{Membership Inference Attack Algorithms} \\
\cline{2-15} 
 & &  & \multicolumn{3}{c}{Metric MIA} &   &   &   \multicolumn{2}{c}{ML-Leaks} & & \multicolumn{4}{c}{BlindMI} \\ 
 \cline{4-6} \cline{9-10} \cline{12-15}
 &LiRA &RMIA &Entropy & \makecell{Modified\\Entropy}& Confidence & Quantile &Merlin & MLLeak1 & MLLeak3 & Shadow & One-class & Diff-w & Diff-single & Diff-bi \\

\Xhline{2\arrayrulewidth}

CIFAR10 & 
\resTwo{59.19}{1.38}{0.74} & \resTwo{56.53}{0.00}{13.06} & \resTwo{65.68}{0.36}{11.93} & \resTwo{66.17}{0.41}{17.03} & \resTwo{66.17}{0.36}{17.10} & \resTwo{61.19}{0.39}{14.66} & \resTwo{56.96}{0.35}{1.39} & 
\resOne{52.90} & \resOne{59.74} & \resOne{54.13} & \resOne{59.30} & \resOne{59.38} & \resOne{59.34} & \resOne{51.88}\\
\hline

\makecell{CIFAR100 \\ Finer-Grained} & 
\resTwo{74.15}{2.50}{1.96} & \resTwo{75.26}{9.44}{47.90} & \resTwo{80.65}{8.47}{55.36} & \resTwo{81.97}{8.92}{60.71} & \resTwo{81.90}{8.68}{60.74} & \resTwo{80.53}{0.24}{47.38} & \resTwo{56.51}{1.53}{0.00} & 
\resOne{73.63} & \resOne{75.05} & \resOne{63.00} & \resOne{68.40} & \resOne{77.95} & \resOne{75.43} & \resOne{67.17}\\
\hline

\makecell{CIFAR100 \\ Superclass} & 
\resTwo{67.56}{0.46}{0.60} & \resTwo{66.21}{2.12}{29.42} & \resTwo{73.65}{0.42}{31.36} & \resTwo{74.67}{0.50}{37.39} & \resTwo{74.64}{0.33}{37.89} & \resTwo{72.53}{0.10}{32.88} & \resTwo{54.31}{0.00}{1.25} & 
\resOne{66.09} & \resOne{68.17} & \resOne{59.12} & \resOne{70.91} & \resOne{69.13} & \resOne{67.56} & \resOne{55.56}\\
\hline

\makecell{CIFAR100 \\ Mislabeled} & 
\resTwo{78.85}{3.26}{2.42} & \resTwo{80.84}{7.22}{58.96} & \resTwo{85.07}{8.15}{66.50} & \resTwo{87.49}{9.18}{72.48} & \resTwo{87.43}{8.94}{72.72} & \resTwo{77.57}{0.24}{0.00} & \resTwo{58.46}{1.77}{10.71} & 
\resOne{80.55} & \resOne{58.63} & \resOne{66.53} & \resOne{82.80} & \resOne{84.77} & \resOne{79.44} & \resOne{72.37}\\
\hline

\makecell{ImageNet100\\(WRN-28)}& 
\resTwo{60.34}{0.36}{0.76} & \resTwo{60.56}{0.68}{0.00} & \resTwo{61.64}{5.30}{7.52} & \resTwo{63.04}{6.35}{11.42} & \resTwo{62.90}{6.24}{11.76} & \resTwo{57.26}{0.32}{0.76} & \resTwo{52.54}{0.00}{0.00} & 
\resOne{50.00} & \resOne{51.06} & \resOne{50.32} & \resOne{53.32} & \resOne{56.76} & \resOne{56.14} & \resOne{54.10}\\

\Xhline{2\arrayrulewidth}
\end{tabular}%
}
\begin{flushleft}
\scriptsize \textit{Note:} In each cell, the top line shows Accuracy, and the bottom line shows (TPR@0.1\%FPR / TNR@0.1\%FNR) Unless otherwise specified, the adopted model is ResNet-18 using the final checkpoint for evaluation. Results on ImageNet-100 are based on WideResNet-28-2 (WRN-28). All evaluated MIA methods are in auditing mode to maximize performance.
\end{flushleft}
\end{table*}

%% file: experiments/architecture.tex
\begin{table*}[h!tbp]
\centering
\tabcolsep=0.18em 
\renewcommand{\arraystretch}{1.25} 
\tiny 
\caption{\small Performance of Different MIA Methods Under Different Architectures on CIFAR100.}

\label{tab:arch_comparison}
\newcommand{\resTwo}[3]{\makecell[c]{#1 \\ \color{gray}\tiny #2~/~#3}}

\newcommand{\resOne}[1]{\makecell[c]{#1 \\ \vphantom{\tiny 0.00~/~0.00} }} 

\resizebox{\textwidth}{!}{%
\begin{tabular}{cccccccccccccccc}
\Xhline{2\arrayrulewidth}

\multirow{3}{*}{\textbf{Architecture}}& \multicolumn{14}{c}{Membership Inference Attack Algorithms} \\
\cline{2-15} 
 & &  & \multicolumn{3}{c}{Metric MIA} &   &   &   \multicolumn{2}{c}{ML-Leaks} & & \multicolumn{4}{c}{BlindMI} \\ 
 \cline{4-6} \cline{9-10} \cline{12-15}
 &LiRA &RMIA &Entropy & \makecell{Modified\\Entropy}& Confidence & Quantile &Merlin & MLLeak1 & MLLeak3 & Shadow & One-class & Diff-w & Diff-single & Diff-bi \\

\Xhline{2\arrayrulewidth}

4-layer CNN & \resTwo{63.66}{0.52}{0.26} & \resTwo{63.44}{0.78}{3.26} & \resTwo{59.59}{2.29}{2.78} & \resTwo{65.22}{3.28}{11.87} & \resTwo{65.05}{3.20}{11.89} & \resTwo{62.27}{0.16}{0.00} & \resTwo{54.37}{0.02}{1.79} &  \resOne{50.00} & \resOne{50.42} & \resOne{51.16} & \resOne{50.12} & \resOne{55.61} & \resOne{53.63} & \resOne{53.57}\\
\hline

VGG-11 &\resTwo{74.48}{1.88}{0.76} & \resTwo{77.29}{10.36}{50.10} & \resTwo{80.30}{7.24}{56.05} & \resTwo{82.03}{7.76}{62.03} & \resTwo{81.97}{7.59}{61.82} & \resTwo{80.77}{0.22}{51.22} & \resTwo{52.90}{0.00}{17.86} & \resOne{73.89} & \resOne{62.32} & \resOne{63.00} & \resOne{71.41} & \resOne{78.91}& \resOne{76.25} & \resOne{66.79}\\
\hline

ResNet-18 & 
\resTwo{74.15}{2.50}{1.96} & \resTwo{75.26}{9.44}{47.90} & \resTwo{80.65}{8.47}{55.36} & \resTwo{81.97}{8.92}{60.71} & \resTwo{81.90}{8.68}{60.74} & \resTwo{80.53}{0.24}{47.38} & \resTwo{56.51}{1.53}{0.00} & 
\resOne{73.63} & \resOne{75.05} & \resOne{63.00} & \resOne{68.40} & \resOne{77.95} & \resOne{75.43} & \resOne{67.17}\\
\hline

ResNet-34 & \resTwo{75.12}{3.14}{2.04} & \resTwo{76.12}{9.92}{50.16} & \resTwo{82.82}{6.68}{63.06} & \resTwo{83.93}{7.08}{66.76} & \resTwo{83.78}{7.05}{66.39} & \resTwo{83.16}{0.66}{56.38} & \resTwo{56.51}{2.14}{0.00} & \resOne{75.69} & \resOne{63.86} & \resOne{63.39} & \resOne{72.46} & \resOne{79.84} & \resOne{77.22} & \resOne{69.54}\\
\hline

ResNet-50 & \resTwo{74.54}{1.26}{2.80} & \resTwo{72.76}{4.32}{44.92} & \resTwo{81.04}{7.46}{56.06} & \resTwo{82.29}{7.97}{61.15} & \resTwo{82.22}{7.82}{61.25} & \resTwo{81.06}{4.20}{50.84} & \resTwo{56.90}{1.39}{3.57} & \resOne{71.24} & \resOne{76.85} & \resOne{62.32} & \resOne{75.45} & \resOne{77.91}& \resOne{74.71} & \resOne{66.46}\\
\hline

Swin-T & \resTwo{64.07}{0.86}{0.66} & \resTwo{62.43}{2.92}{22.08} & \resTwo{66.91}{2.81}{27.43} & \resTwo{67.93}{2.36}{30.74} & \resTwo{67.84}{2.16}{30.72} & \resTwo{65.16}{0.18}{21.88} & \resTwo{56.52}{1.66}{0.00} & \resOne{56.31} & \resOne{57.59} & \resOne{55.19} & \resOne{61.27} & \resOne{61.89} & \resOne{63.04} & \resOne{53.40}\\

\Xhline{2\arrayrulewidth}
\end{tabular}%
}
\begin{flushleft}
\scriptsize \textit{Note:} In each cell, the top line shows Accuracy, and the bottom line shows (TPR@0.1\%FPR / TNR@0.1\%FNR). Unless otherwise specified, models are trained from scratch on CIFAR-100, and the final checkpoint is used for MIA evaluation. For Swin-T, parameters are initialized with an ImageNet pre-trained checkpoint due to poor performance when training from scratch. All evaluated MIA methods are in auditing mode to maximize performance.
\end{flushleft}
\end{table*}

%% file: experiments/train_alg.tex
\begin{table*}[h!tbp]
    \centering
    \tabcolsep=0.18em
    \renewcommand{\arraystretch}{1.25}
    \tiny
     \caption{\small Performance of different MIA methods under different training algorithms. }
    \label{tab:train_alg}
    \newcommand{\resTwo}[3]{\makecell[c]{#1 \\ \color{gray}\tiny #2~/~#3}}
    \newcommand{\resOne}[1]{\makecell[c]{#1 \\ \vphantom{\tiny 0.00~/~0.00} }}
    \resizebox{\textwidth}{!}{%
    \begin{tabular}{cccccccccccccccc}
    \Xhline{2\arrayrulewidth}
    \multirow{3}{*}{\textbf{Method}} & \multicolumn{14}{c}{Membership Inference Attack Algorithms} \\
    \cline{2-15}
     & &  & \multicolumn{3}{c}{Metric MIA} &   &   &   \multicolumn{2}{c}{ML-Leaks} & & \multicolumn{4}{c}{BlindMI} \\
    \cline{4-6} \cline{9-10} \cline{12-15}
     & LiRA & RMIA & Entropy & \makecell{Modified\\Entropy} & Confidence & Quantile & Merlin & MLLeak1 & MLLeak3 & Shadow & One-class & Diff-w & Diff-single & Diff-bi \\
    \Xhline{2\arrayrulewidth}
    \multicolumn{15}{l}{\textit{\textbf{Training Algorithm, DP-SGD}}} \\ \hline
    SGD & 
    \resTwo{74.15}{2.50}{1.96} & \resTwo{75.26}{9.44}{47.90} & \resTwo{80.65}{8.47}{55.36} & \resTwo{81.97}{8.92}{60.71} & \resTwo{81.90}{8.68}{60.74} & \resTwo{80.53}{0.24}{47.38} & \resTwo{56.51}{1.53}{0.00} & 
    \resOne{73.63} & \resOne{75.05} & \resOne{63.00} & \resOne{68.40} & \resOne{77.95} & \resOne{75.43} & \resOne{67.17}\\
    \hline

    DP-SGD  & \resTwo{52.01}{0.00}{0.04} & \resTwo{66.73}{0.14}{3.50} & \resTwo{66.59}{2.70}{9.89} & \resTwo{70.34}{3.04}{26.92} & \resTwo{70.29}{3.06}{28.00} & \resTwo{69.89}{0.30}{0.00} & \resTwo{57.05}{1.51}{7.14} & \resOne{54.32} & \resOne{63.93} & \resOne{63.27} & \resOne{64.03} & \resOne{67.49} & \resOne{59.75} & \resOne{53.90}\\
    \hline
    
    \multicolumn{15}{l}{\textit{\textbf{Post Training, Fine-tune, CIFAR-100}}} \\ \hline

From Scratch & \resTwo{74.54}{1.26}{2.80} & \resTwo{72.76}{4.32}{44.92} & \resTwo{81.04}{7.46}{56.06} & \resTwo{82.29}{7.97}{61.15} & \resTwo{82.22}{7.82}{61.25} & \resTwo{81.06}{4.20}{50.84} & \resTwo{56.90}{1.39}{3.57} & \resOne{71.24} & \resOne{76.85} & \resOne{62.32} & \resOne{75.45} & \resOne{77.91}& \resOne{74.71} & \resOne{66.46} \\
\hline
    \makecell{Pretrain \\ + Finetune} & \resTwo{59.93}{0.22}{0.20} & \resTwo{65.90}{0.00}{25.64} & \resTwo{57.14}{2.05}{2.05} & \resTwo{58.20}{1.05}{2.28} & \resTwo{58.24}{0.91}{2.59} & \resTwo{72.56}{0.30}{30.50} & \resTwo{56.56}{1.81}{0.00} & \resOne{50.00} & \resOne{50.69} & \resOne{50.10} & \resOne{51.12} & \resOne{50.74} & \resOne{50.62} & \resOne{50.45}\\   
    \hline

    \multicolumn{15}{l}{\textit{\textbf{Post Training, Fine-tune, CIFAR10}}} \\ \hline
  From Scratch &  \resTwo{57.60}{0.96}{0.18} &  \resTwo{58.14}{0.00}{10.82} & 
  \resTwo{59.47}{0.08}{6.68} & \resTwo{60.27}{0.06}{12.52} & \resTwo{60.20}{0.06}{12.48} & \resTwo{60.05}{0.16}{9.60} & \resTwo{57.04}{0.20}{0.41} & \resOne{50.00} & \resOne{58.85} & \resOne{53.36} & \resOne{56.38} & \resOne{57.74} & \resOne{58.49} & \resOne{51.80}\\
  \hline
  \makecell{Pretrain \\ + Finetune} & \resTwo{58.12}{0.24}{0.30} &  \resTwo{56.58}{0.00}{7.34} & \resTwo{51.12}{0.10}{0.16} & \resTwo{51.24}{0.00}{0.08} & \resTwo{51.20}{0.00}{0.08} & \resTwo{58.69}{0.09}{7.86} & \resTwo{56.91}{0.27}{0.00} & \resOne{50.73} & \resOne{50.13} & \resOne{51.27} & \resOne{50.35} & \resOne{51.15} & \resOne{50.84} & \resOne{50.38}\\
    \Xhline{2\arrayrulewidth}
    \end{tabular}%
    }

\begin{flushleft}
\scriptsize \textit{Note:} In each cell, the top line shows Accuracy, and the bottom line shows (TPR@0.1\%FPR / TNR@0.1\%FNR). For SGD and DP-SGD, the adopted model is ResNet-18. For post training, the adopted model is ResNet-50. The evaluated MIA methods are in auditing mode to maximize their performance.
\end{flushleft}
\end{table*}

%% file: experiments/unlearn_result_reduced.tex
\noindent \textbf{Machine Unlearning.} Machine unlearning aims to remove the effect of a forget set 
$\mathcal D_{\mathrm{forget}} \subset \mathcal D_{\mathrm{train}}$ 
from a pretrained model while preserving performance on the retained set 
$\mathcal D_{\mathrm{retain}} = \mathcal D_{\mathrm{train}} \setminus \mathcal D_{\mathrm{forget}}$. 
The standard reference is a retrained model trained only on 
$\mathcal D_{\mathrm{retain}}$, where samples in $\mathcal D_{\mathrm{forget}}$ 
should behave as non-members. An ideal unlearned model should satisfy two properties: 
it should remove membership signals of $\mathcal D_{\mathrm{forget}}$, and its output distribution 
should be close to that of the retrained model. We denote membership status of different models and datasets in Figure~\ref{fig:mu}. 

The unlearning goal gives rise to two MIA-based auditing axes: \emph{fixed-data} and \emph{fixed-model} audit. 
In \emph{fixed-data} audit, MIAs are trained on $\gD_{\mathrm{forget}}$, which is considered members for the pretrained model and non-members for the retrained model.
In \emph{fixed-model} audit, MIAs are trained on the retrained model, for which $\gD_{\mathrm{retain}}$ is considered members and $\gD_{\mathrm{test}}$ is considered non-members. We evaluate existing MIAs along these two axes on two popular unlearning methods, SalUn~\cite{fan2023salun} and SFRon~\cite{huang2024unified}, under four CIFAR100 forgetting settings: removing one category, and removing 5\%, 10\%, or 50\% of randomly sampled training data with balanced categories.

In fixed-data audit, a reliable MIA should identify $\gD_{\mathrm{forget}}$ as member data for the pretrained model but non-member data for the retrained model. 
Fixed-data auditing is independent of the unlearning algorithm, as it does not evaluate the unlearned model. Instead, it aims to determine the membership status of a fixed dataset across different models. This contrasts with the previously investigated scenarios, which assess the membership status of different datasets on a single model.
Table~\ref{tab:ars_fmf} presents the evaluation results, detailing the overall accuracy, true positive rate, and true negative rate. Our findings indicate that BlindMI One-class generally achieves the best performance, although LiRA slightly outperforms it in the single-category unlearning scenario. Note that MetricMIA and Shadow are inapplicable to this specific setting, as their methodologies rely on per-class information.

\begin{figure}[htbp]
    \centering
    \includegraphics[width=0.97\linewidth]{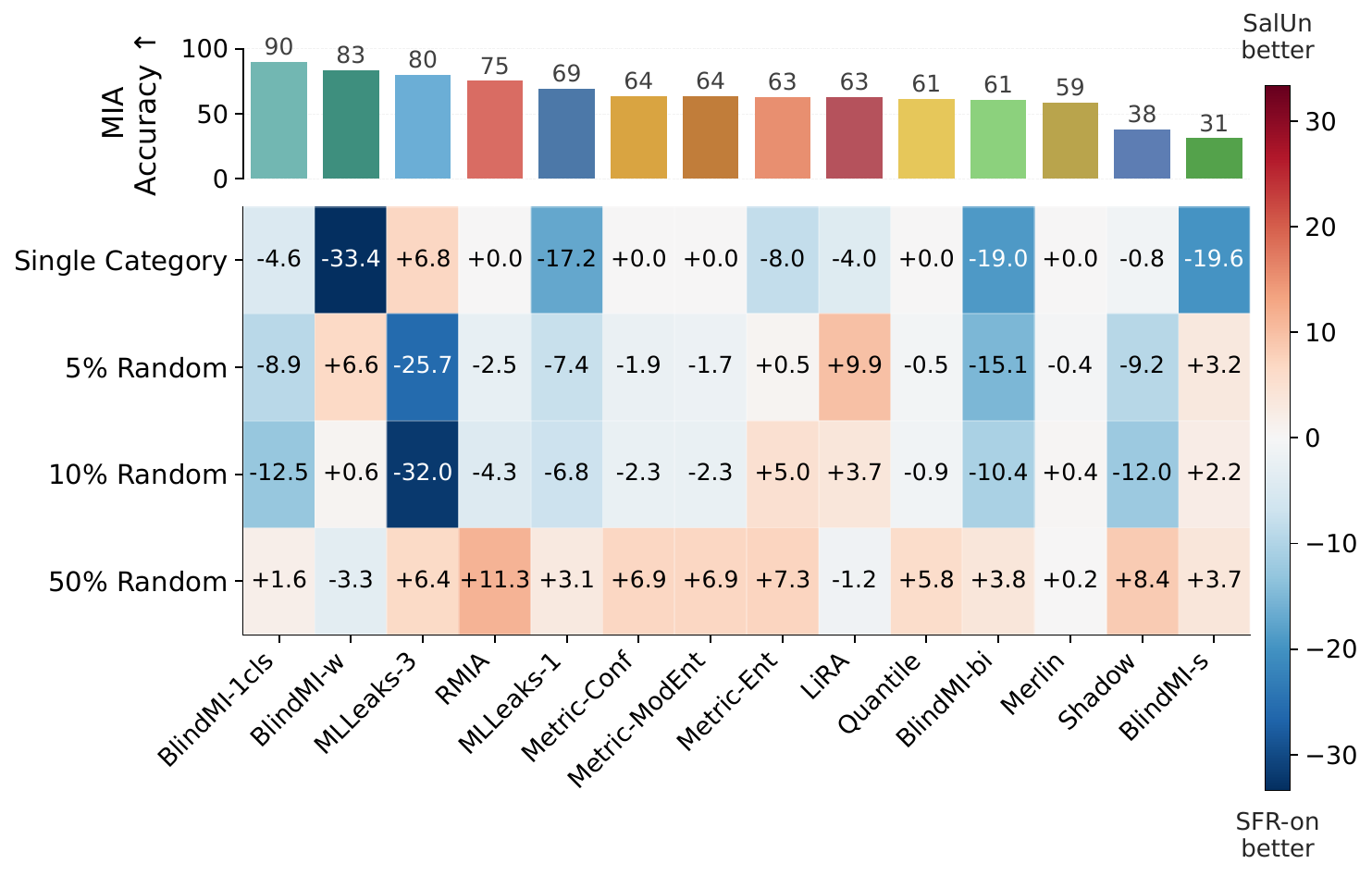}
    \vspace{-0.6em}
    \caption{\small Comparison between SalUn and SFR-on under fixed-model audit. 
    The top bar shows the performance for each MIA. The MIA methods are sorted in the descending order of their performance. The heatmap shows $\Delta\mathrm{OptIndis}=\mathrm{OptIndis}_{\mathrm{SFR\text{-}on}}-\mathrm{OptIndis}_{\mathrm{SalUn}}$. Positive and negative values indicate the advantages of SalUn and SFR-on under the corresponding MIA, respectively.}
    \label{fig:mu_mia_conditioned_heatmap}
    \vspace{-0.8em}
\end{figure}

Most unlearning literature~\cite{fan2023salun, huang2024unified} adopts the fixed-model audit and evaluate the unlearning methods by the output-indistinguishability, which is the accuracy discrepancy of the MIA methods between the unlearned model and the retrained model.
Specifically, we train each MIA method on the retrained model by treating $\gD_{\mathrm{retain}}$ as members and $\gD_{\mathrm{test}}$ as non-members, and then apply the same MIA attacker to different unlearned models. We visualize $\Delta \mathrm{OptIndis} = \mathrm{OptIndis}_{\mathrm{SFR\text{-}on}} -\mathrm{OptIndis}_{\mathrm{SalUn}}$ and the accuracy of different MIAs in Figure~\ref{fig:mu_mia_conditioned_heatmap}, where $\mathrm{OptIndis}_{\mathrm{SFR\text{-}on}}$ and $\mathrm{OptIndis}_{\mathrm{SalUn}}$ indicate the accuracy discrepancy for SFR-on and SalUn, respectively.
Therefore, negative values in Figure~\ref{fig:mu_mia_conditioned_heatmap} indicate that SFR-on is closer to the retrained model and thus better, whereas a positive value indicates that SalUn is better. Generally, under top-ranking MIAs, SFR-on is better in the single-category, $5\%$ random, and $10\%$ random settings, while SalUn is better in the $50\%$ random setting.
Nevertheless, the sign and magnitude of $\Delta \mathrm{OptIndis}$ vary across MIAs. This result raises the attention that the choice of MIA would change the ranking between unlearning methods, so we should choose appropriate MIA methods before evaluating unlearning algorithms.

\input{experiments/unlearn_fmf}

%% file: experiments/unlearn_fmf.tex
\begin{table*}[h!tbp]
    \centering
    \tabcolsep=0.18em
    \renewcommand{\arraystretch}{1.25}
    \tiny
    \caption{\small Performance of Different MIA methods under Fixed-Data Audit in Machine Unlearning.}
    \label{tab:ars_fmf}
    \newcommand{\resTwo}[3]{\makecell[c]{#1 \\ \color{gray}\tiny #2~/~#3}}
    \newcommand{\resOne}[1]{\makecell[c]{#1 \\ \vphantom{\tiny 0.0~/~0.0}}}
    \newcommand{\best}[1]{\textbf{#1}}
    \newcommand{\second}[1]{\uline{#1}}
    \resizebox{\textwidth}{!}{%
    \begin{tabular}{ccccccccccccccc}
    \Xhline{2\arrayrulewidth}
    \multirow{3}{*}{\textbf{Setting}} & \multicolumn{14}{c}{Membership Inference Attack Algorithms} \\
    \cline{2-15}
     & & & \multicolumn{3}{c}{Metric MIA} & & & \multicolumn{2}{c}{ML-Leaks} & & \multicolumn{4}{c}{BlindMI} \\
    \cline{4-6} \cline{9-10} \cline{12-15}
     & LiRA & RMIA & Entropy & \makecell{Modified\\Entropy} & Confidence & Quantile & Merlin & MLLeak1 & MLLeak3 & Shadow & One-class & Diff-w & Diff-single & Diff-bi \\
    \Xhline{2\arrayrulewidth}

    \textit{\textbf{Single Category}} 
    & \resTwo{\best{94.4}}{90.0}{98.8} 
    & \resTwo{76.4}{52.8}{100.0} 
    & \resTwo{--}{--}{--} 
    & \resTwo{--}{--}{--}  
    & \resTwo{--}{--}{--} 
    & \resTwo{39.6}{0.4}{78.8} 
    & \resTwo{87.0}{74.0}{100.0} 
    & \resTwo{65.4}{100.0}{30.8} 
    & \resTwo{85.0}{78.0}{92.0} 
    & \resTwo{--}{--}{--}  
    & \resTwo{91.2}{100.0}{82.4} 
    & \resTwo{\second{91.8}}{100.0}{83.6} 
    & \resTwo{51.0}{86.8}{15.2} 
    & \resTwo{62.2}{65.2}{59.2} \\
    \hline

    \textit{\textbf{5\% Random}} 
    & \resTwo{64.0}{97.4}{30.5} 
    & \resTwo{77.4}{78.9}{75.8} 
    & \resTwo{84.0}{96.1}{71.9} 
    & \resTwo{84.8}{96.9}{72.7} 
    & \resTwo{\second{84.9}}{96.7}{73.0} 
    & \resTwo{81.8}{97.5}{66.1} 
    & \resTwo{49.9}{46.4}{53.4} 
    & \resTwo{81.8}{99.7}{63.8} 
    & \resTwo{81.0}{79.0}{82.9} 
    & \resTwo{65.6}{100.0}{31.1} 
    & \resTwo{\best{88.3}}{100.0}{76.6} 
    & \resTwo{84.1}{99.9}{68.2} 
    & \resTwo{52.6}{84.9}{20.3} 
    & \resTwo{60.5}{65.2}{55.7} \\
    \hline

    \textit{\textbf{10\% Random}} 
    & \resTwo{63.7}{96.6}{30.8} 
    & \resTwo{77.8}{79.7}{75.9} 
    & \resTwo{83.4}{95.8}{71.0} 
    & \resTwo{84.4}{96.8}{71.9} 
    & \resTwo{84.5}{96.7}{72.3} 
    & \resTwo{81.6}{97.3}{65.9} 
    & \resTwo{50.5}{46.7}{54.2} 
    & \resTwo{77.9}{99.9}{55.8} 
    & \resTwo{\second{86.9}}{76.4}{97.3} 
    & \resTwo{65.0}{100.0}{29.9} 
    & \resTwo{\best{89.2}}{99.9}{78.5} 
    & \resTwo{83.4}{99.8}{66.9} 
    & \resTwo{53.2}{84.5}{21.8} 
    & \resTwo{60.5}{65.0}{56.0} \\
    \hline

    \textit{\textbf{50\% Random}} 
    & \resTwo{72.5}{96.5}{48.4} 
    & \resTwo{77.0}{78.8}{75.1} 
    & \resTwo{88.8}{95.7}{81.9} 
    & \resTwo{89.8}{96.4}{83.1} 
    & \resTwo{\second{89.9}}{96.2}{83.6} 
    & \resTwo{88.0}{98.3}{77.7} 
    & \resTwo{49.7}{46.4}{53.0} 
    & \resTwo{84.1}{99.9}{68.2} 
    & \resTwo{70.2}{76.1}{64.3} 
    & \resTwo{70.4}{100.0}{40.8} 
    & \resTwo{\best{95.2}}{99.9}{90.5} 
    & \resTwo{84.5}{99.9}{69.0} 
    & \resTwo{52.6}{85.7}{19.4} 
    & \resTwo{61.2}{63.9}{58.5} \\
    \Xhline{2\arrayrulewidth}

    \textbf{Average} 
    & \resTwo{73.6}{95.1}{52.1} 
    & \resTwo{77.1}{72.5}{81.7} 
    & \resTwo{85.4}{95.9}{74.9} 
    & \resTwo{86.4}{96.7}{75.9} 
    & \resTwo{86.4}{96.5}{76.3} 
    & \resTwo{72.8}{73.4}{72.1} 
    & \resTwo{59.3}{53.4}{65.1} 
    & \resTwo{77.3}{99.9}{54.7} 
    & \resTwo{80.8}{77.4}{84.1} 
    & \resTwo{67.0}{100.0}{33.9} 
    & \resTwo{\best{91.0}}{100.0}{82.0} 
    & \resTwo{\second{85.9}}{99.9}{71.9} 
    & \resTwo{52.4}{85.5}{19.2} 
    & \resTwo{61.1}{64.8}{57.4} \\
    \Xhline{2\arrayrulewidth}
    \end{tabular}
    }
\begin{flushleft}
\scriptsize \textit{Note:}  
In each cell, the top line shows Accuracy, and the bottom line shows (true positive rate / true negative rate).
The best results in each row are shown in bold.
MetricMIA and ShadowMIA are excluded from single category unlearning since their per-class calibration is invalid for class unlearning: the forget class is absent from shadow calibration sets. 
\end{flushleft}
\end{table*}

%% file: Sections/conclusion.tex
\section{Discussion}

\subsection{Comparison among Top-Performing MIA Methods}

% Our preceding evaluations across diverse pipeline configurations (excluding the machine unlearning scenario) consistently demonstrate that RMIA, Metric MIA, and Quantile MIA MIA outperform other baseline methods. Consequently, in this section, we further explore the comparative sensitivity and operational boundaries of these state-of-the-art methods by analyzing their Detection Error Tradeoff (DET) curves in Figure \ref{fig:two_rows_grid}, and we have following conclusions:

\begin{figure}[!t]
\label{fig:det}
    \centering
    % === 第一行 (Row 1) ===
    \subfigure[ResNet18 on CIFAR100]{
        \includegraphics[width=0.3\textwidth]{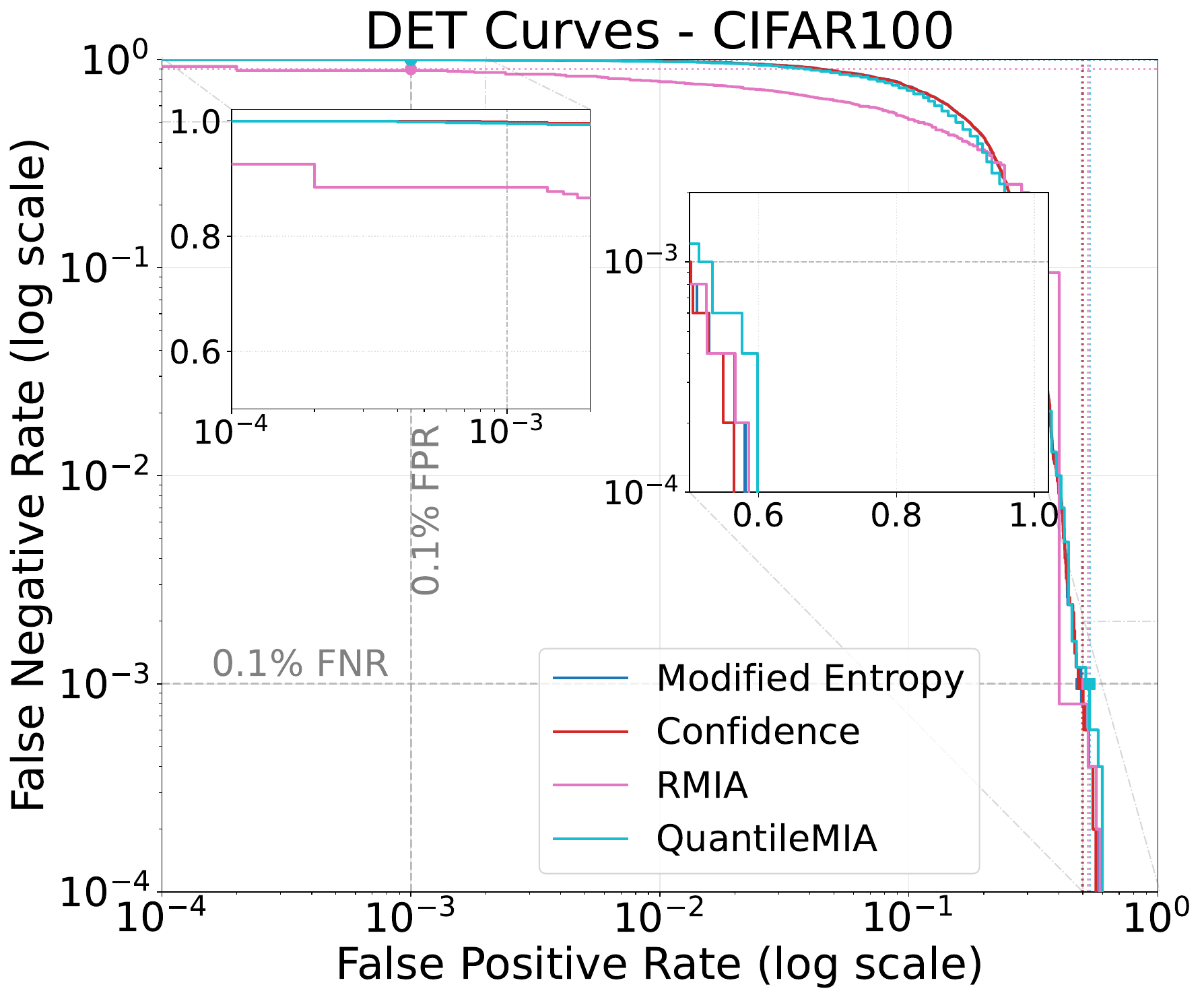}
    }
    \hfill % 使用 \hfill 自动填充横向间距，使三张图均匀分布
    \subfigure[Swin-T on CIFAR100]{
        \includegraphics[width=0.3\textwidth]{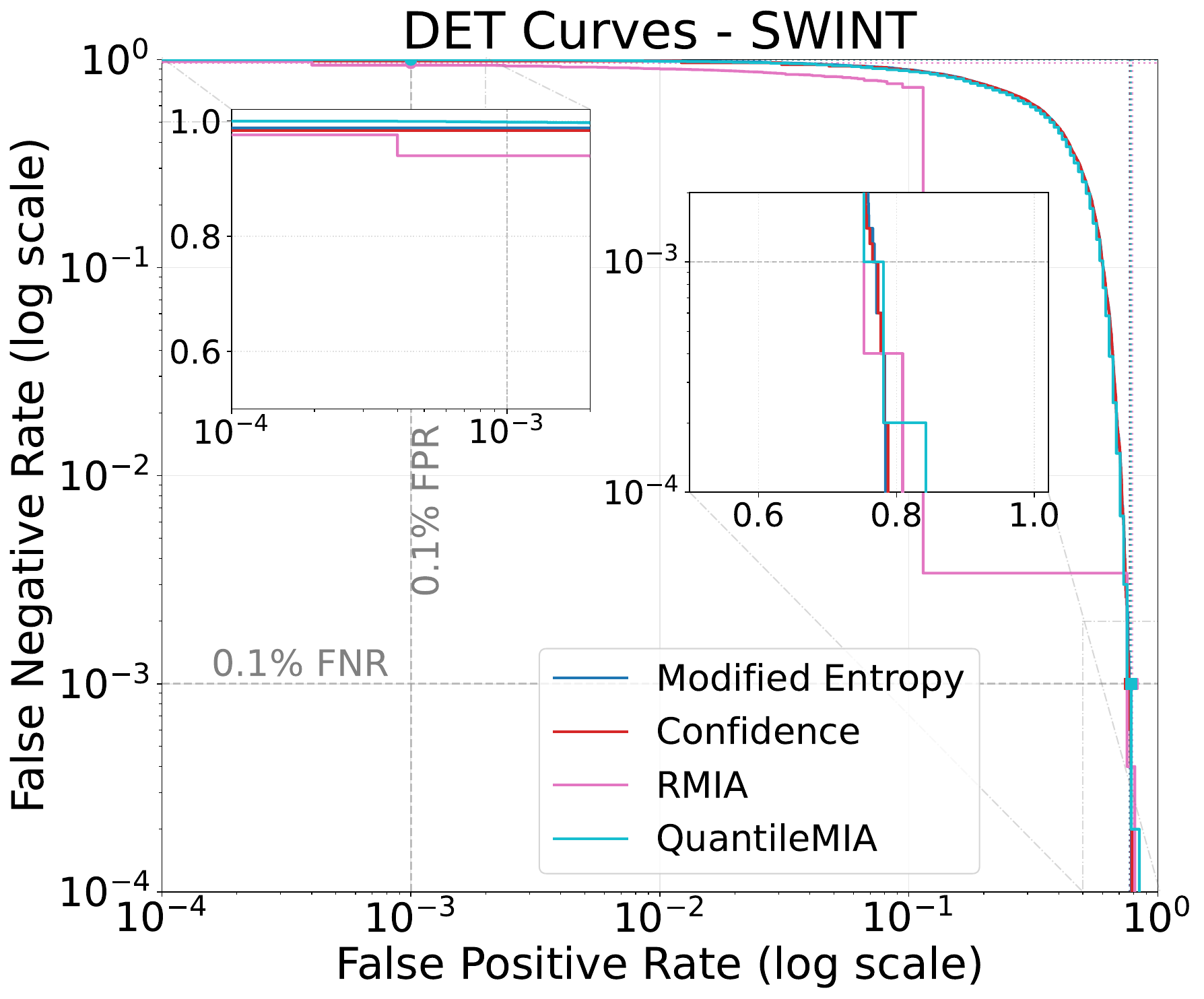}
    }
    \hfill
    \subfigure[4-CNN on CIFAR100]{
        \includegraphics[width=0.3\textwidth]{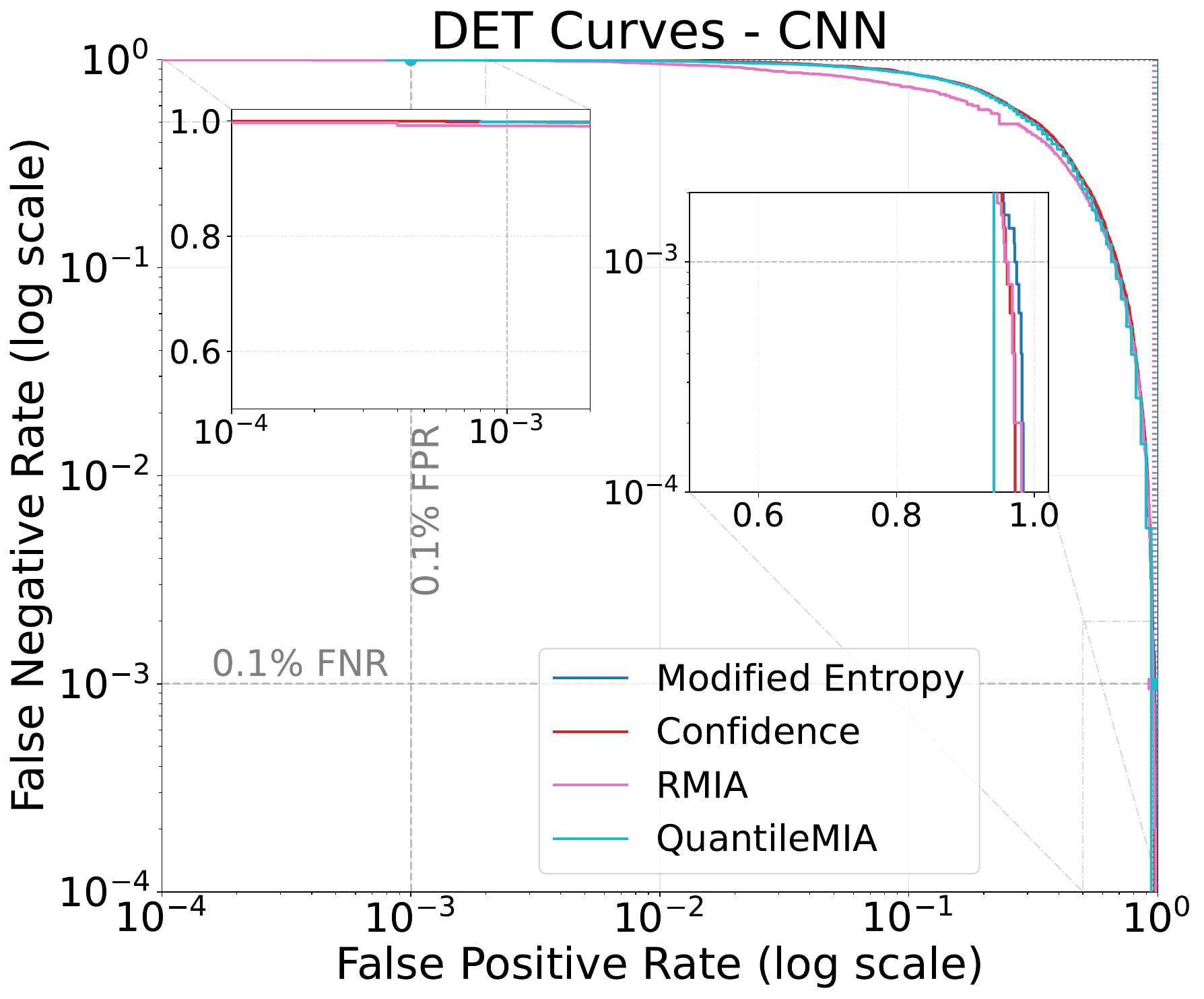}
      
    }

    \vspace{1em} % 增加上下两行之间的垂直间距

    % === 第二行 (Row 2) ===
    \subfigure[ResNet-18 on CIFAR10]{
        \includegraphics[width=0.3\textwidth]{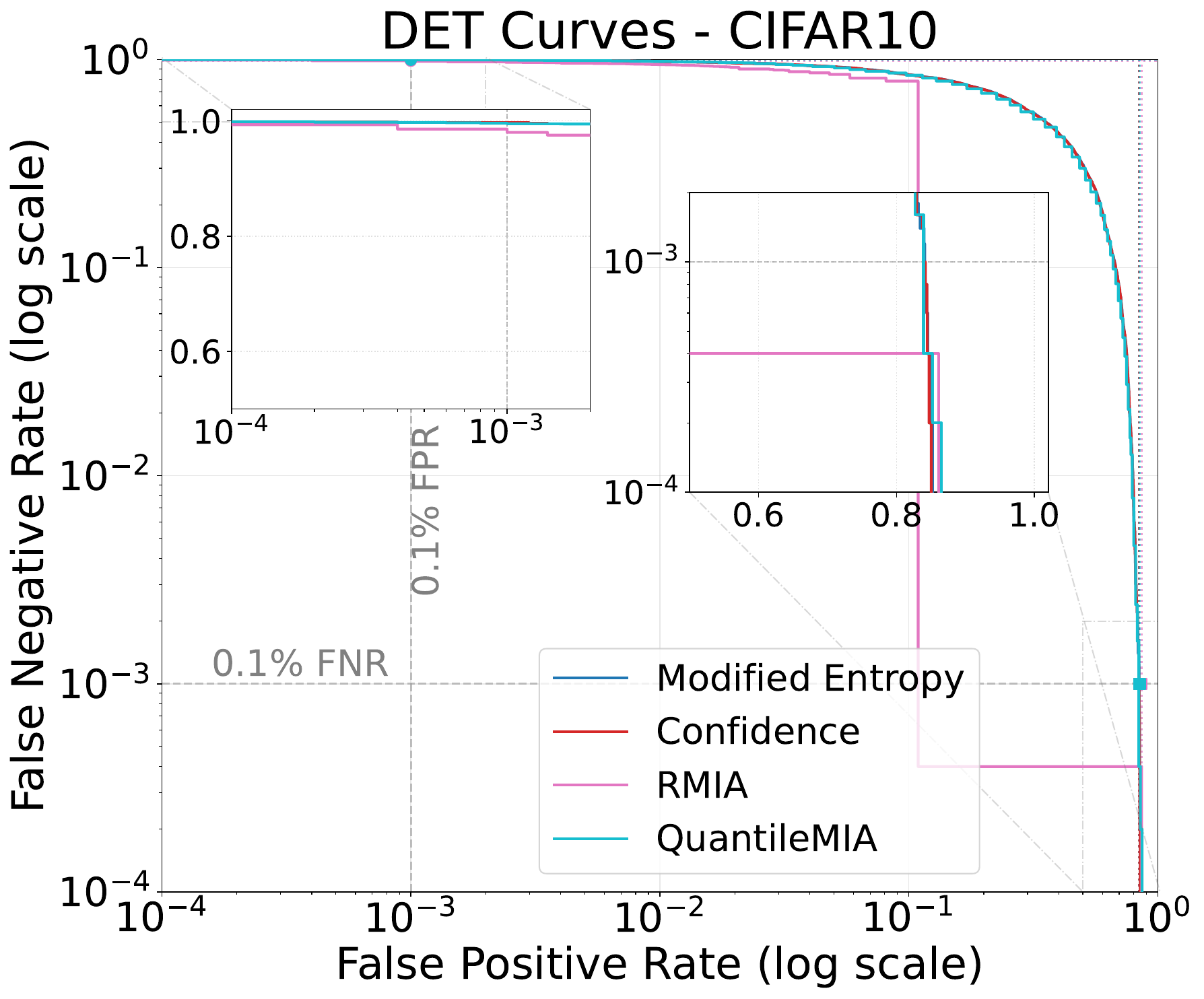}
    }
    \hfill
    \subfigure[WRN-28 on ImageNet100]{
        \includegraphics[width=0.3\textwidth]{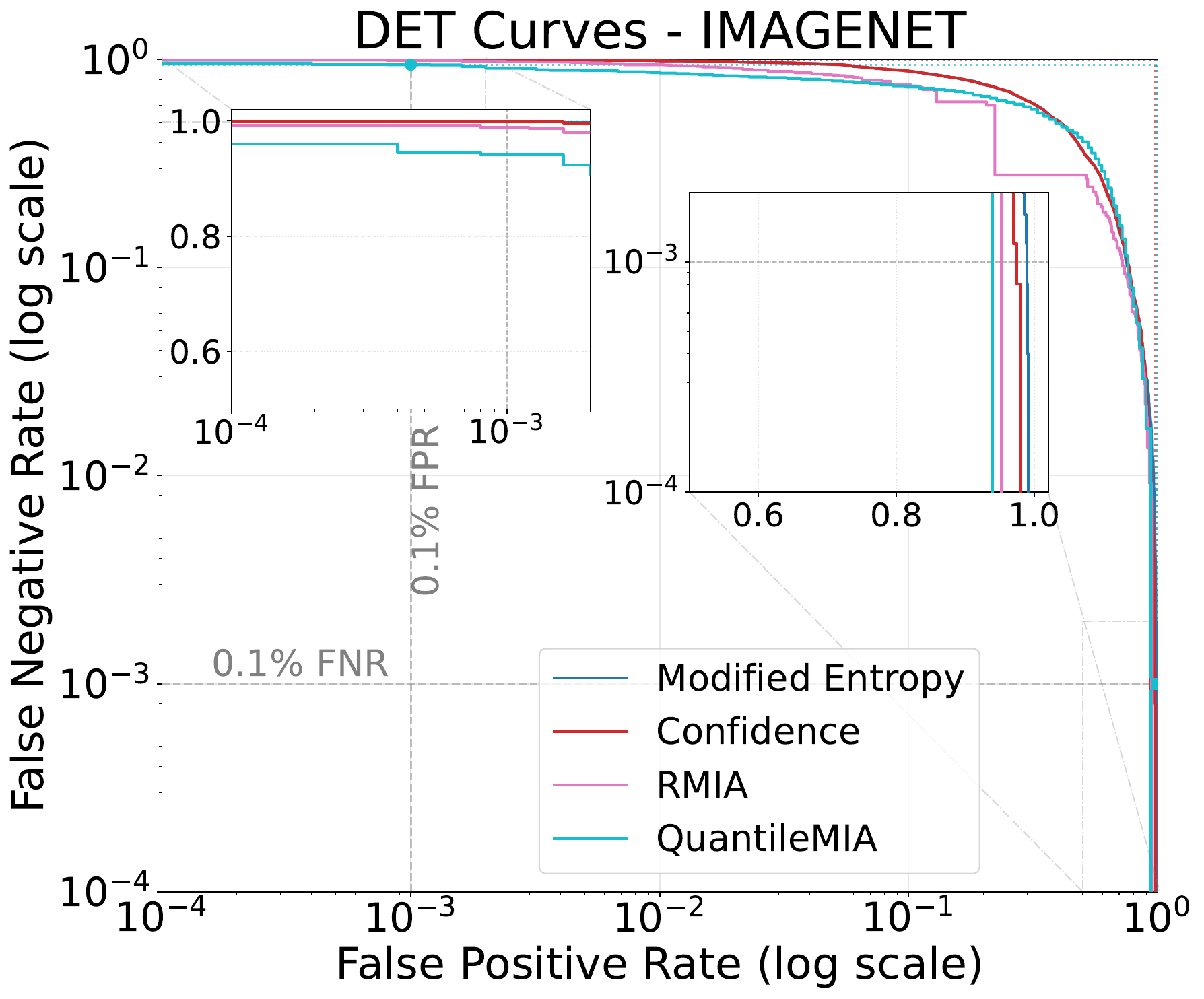}
    }
    \hfill
    \subfigure[ResNet-18 on CIFAR100 with DPSGD]{
        \includegraphics[width=0.3\textwidth]{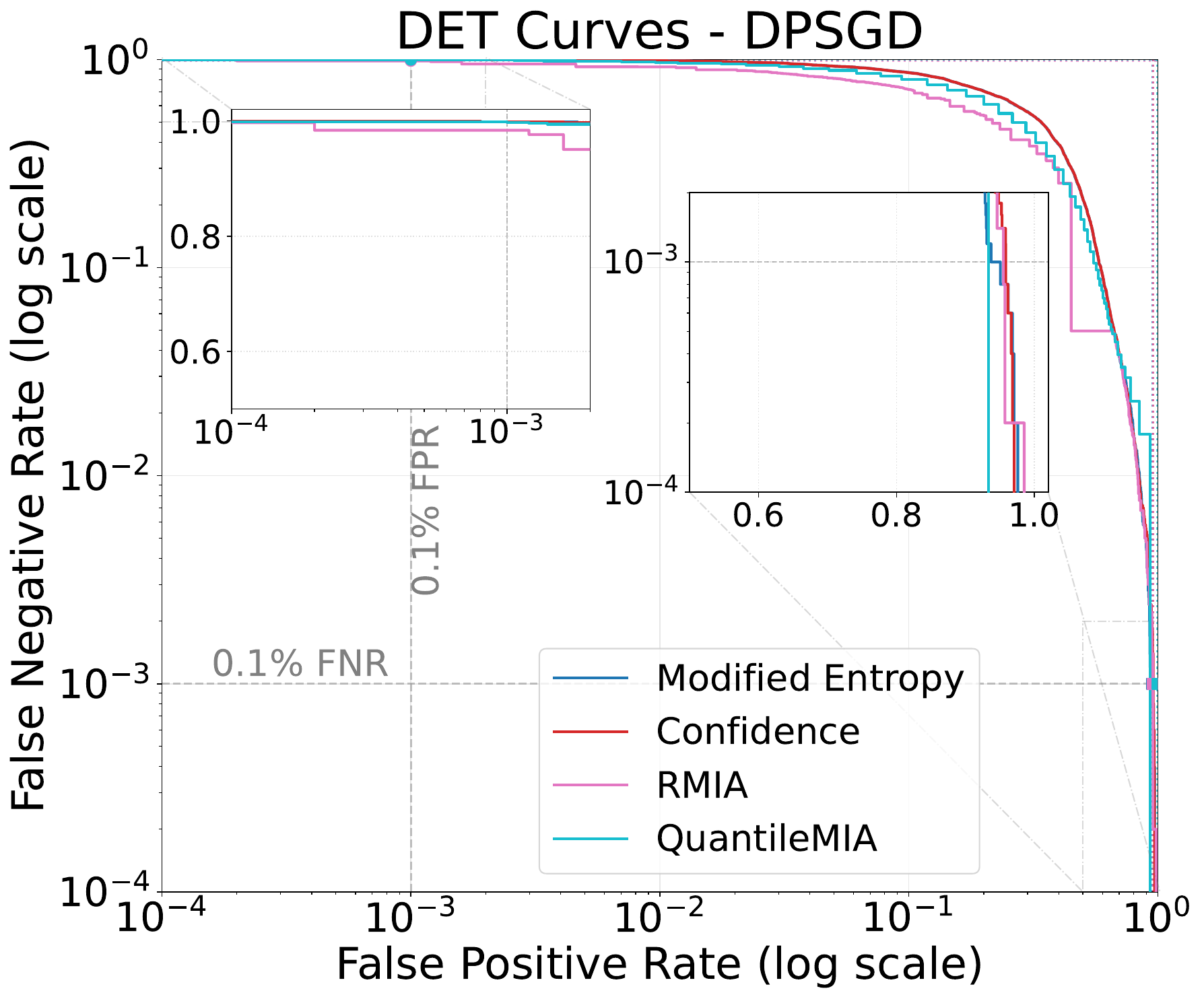}
    }
    \caption{Comparative analysis of DET curves across six pipeline settings. Top row: CIFAR100 dataset under different architecture. Bottom row: Different dataset and training algorithms under ResNet(Results on ImageNet-100 are based on WideResNet-28-2).All methods are evaluated in auditing mode.}
    \label{fig:two_rows_grid}
\end{figure}

Our full-pipeline evaluations in the last section have identified RMIA, Metric MIA and Quantile MIA as top-performing methods in various scenarios, so we further explore the comparative sensitivity and operational boundaries of them.
Specifically, as all these MIAs have hyper parameters, we analyze their Detection Error Tradeoff (DET) curves, i.e., FPR-FNR tradeoff under various hyper parameters, in Figure~\ref{fig:two_rows_grid}. Our discoveries on the results in Section~\ref{sec:exp_results} and Figure~\ref{fig:two_rows_grid} are summarized as follows:

\noindent \textbf{Metric MIA: Strong but Instable Performance.} Metric MIA demonstrates superior overall performance under the auditing mode on various datasets and model architectures (Table~\ref{tab:data_complexity_comparison}, Table~\ref{tab:arch_comparison}). However, it suffers considerable performance degradation in some challenging scenarios. Specifically, its performance drops in the attack mode (Table~\ref{tab:gap_sensitivity_with_lines}) and during the finetuning phase (Table~\ref{tab:train_alg}). These observations indicate that Metric MIA is sensitive to its hyper parameters (i.e., threshold) and struggles to capture training instances with smaller parameter gradients or lower learning rates.

% While Metric MIA variants demonstrate superior overall performance across most standard metrics, they are the most vulnerable to pipeline variations such as model capacity expansion or regularization changes. Because Metric MIA directly utilizes raw output logits for fine-grained, label-wise analysis, it excels at capturing minute confidence differences. However, this direct reliance on absolute output magnitudes renders it highly susceptible to numerical scaling perturbations induced by the underlying model architecture.

\noindent \textbf{Quantile MIA: Stable Performance.} Quantile MIA generally serves as a complementary approach to Metric MIA; while its baseline performance is slightly lower across datasets and architectures in standard training (Table~\ref{tab:data_complexity_comparison}, Table~\ref{tab:arch_comparison}), it demonstrates remarkable resilience in more challenging scenarios. Specifically, Quantile MIA maintains high efficacy in the attack mode (Table~\ref{tab:gap_sensitivity_with_lines}), as well as in DP-SGD or finetuning (Table~\ref{tab:train_alg}). These results indicate that Quantile MIA is a robust methodology, characterized by stable performance despite variations in hyper parameters and training algorithms.

% Quantile MIA MIA is highly resilient to data distribution shifts and degraded auxiliary data quality. This performance stability stems from its reliance on regression models to calibrate sample difficulty. The regression formulation provides an intrinsic smoothing effect, significantly mitigating the impact of environmental perturbations and non-ideal operational conditions.

\noindent \textbf{RMIA: Asymmetric Superiority in Low-FPR Regimes.} RMIA excels in extreme, high-confidence leakage scenarios (achieving high TPR at a strictly low FPR), but underperforms in high-recall screening tasks (TNR at low FNR). Its core mechanism—the likelihood ratio test—drastically amplifies the signal of a few memorized samples. However, this inherently causes a sparse, polarized probability distribution, lacking the smooth discriminability required to excel across broader, global evaluation metrics.

Furthermore, while the previously discussed MIA methods excel at distinguishing between different datasets on a single model, Section~\ref{subsec:post-train} and Table~\ref{tab:ars_fmf} demonstrate that \textbf{BlindMI One-class} achieves the highest performance when evaluating a single dataset across multiple models. We omit the DET curves for BlindMI One-class because it does not rely on threshold hyper parameters.

\subsection{Overfitting leads to MIA vulnerability}
\begin{figure}[ht]
\vspace{-0.5em}
    \centering
    \subfigure[Dataset]{
    \includegraphics[width=0.31\textwidth]{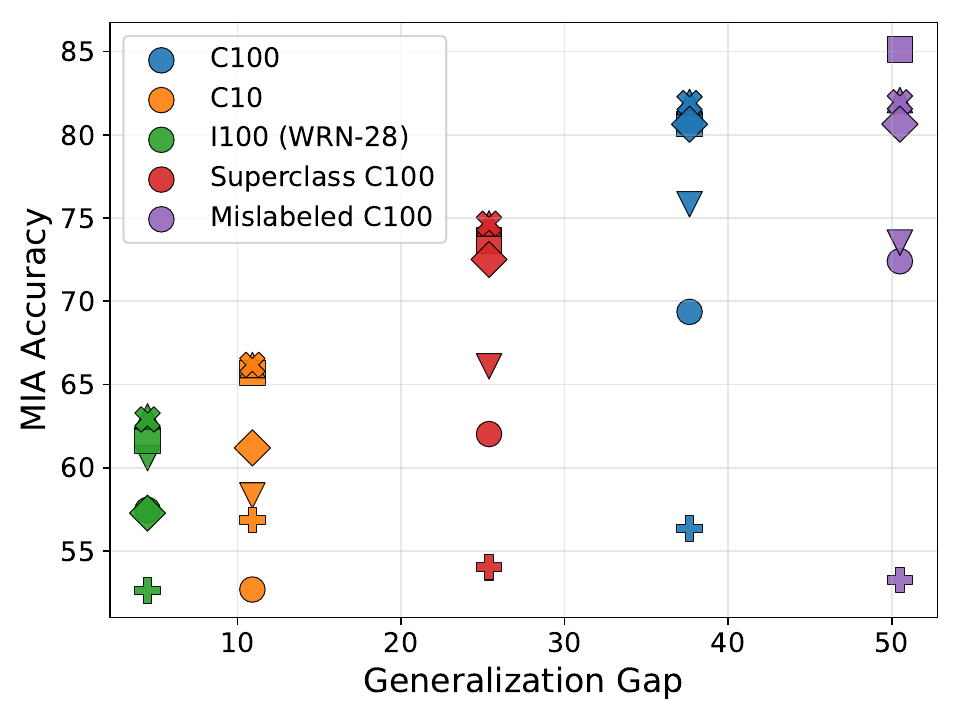}
    }
    \subfigure[Architecture]{
    \includegraphics[width=0.31\textwidth]{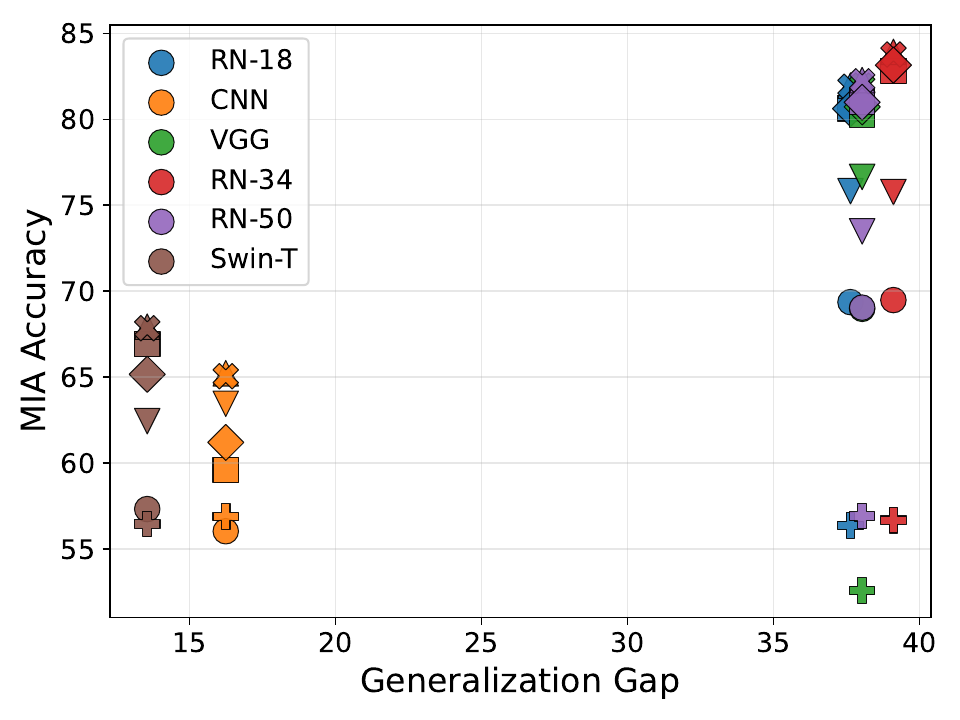}
    }
    \subfigure[Training Algorithm]{
    \includegraphics[width=0.31\textwidth]{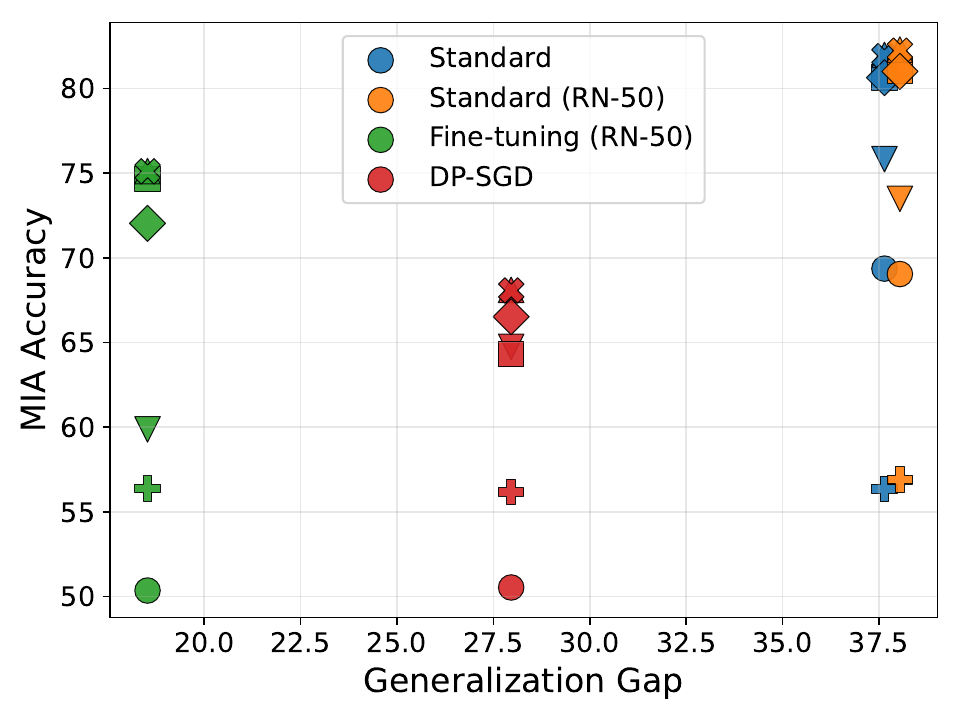}
    }
    \vspace{-0.5em}
    \caption{\small Relationship between MIA accuracy and the degree of overfitting. The analysis is conducted from different aspects, e.g., \textbf{(a)} different datasets, ``C" denotes CIFAR, ``I" denotes ImageNet, \textbf{(b)} different architectures, and \textbf{(c)} different training algorithms. The x-axis is the generalization gap. A larger generalization gap indicates severer overfitting. The y-axis is the accuracy of MIA methods. Different colors represent different settings. Different markers represent different MIA methods. LiRA (\mycirc), RMIA (\mydowntri), Entropy (\mysquare), Modified Entropy (\myuptri), Confidence (\myx), Quantile MIA (\mydiamond), and Merlin (\myplus) in auditing mode are tested here.}
    \label{fig:mia_vs_acc}
    \vspace{-1em}
\end{figure}

We illustrate the relationship between MIA accuracy and the degree of overfitting in Figure \ref{fig:mia_vs_acc}. The analysis includes different aspects, e.g., different datasets, architectures, and training algorithms, with the same settings as those in Appendix \ref{subsec:app_mia_scratch}.
Note that the degree of overfitting is indicated by the generalization gap, i.e., the gap between the training and the test accuracy.
Detailed performance statistics of different settings are summarized in Table \ref{tab:performance_summary} of Appendix \ref{subsec:app_mia_scratch}.

As depicted in Figure \ref{fig:mia_vs_acc}, the accuracy of most MIA methods exhibits a general upward trend as the generalization gap of target models increases. 
In  Figure \ref{fig:mia_curve_mis} and Figure~\ref{fig:train_alg_mia_ep}, we also plot the generalization gap and observe that it is strongly correlated with the MIA performance.
Therefore, many of the observations in Section~\ref{sec:exp_results}, including analyses across different datasets, models, and algorithms, can be understood through the lens of overfitting. Specifically, overfitting plays a significant role in increasing the model's vulnerability to MIAs.

The underlying mechanism for this association can be attributed to the working principle of existing MIA methods: these attacks determine the membership of a given data sample by leveraging discrepancies in the output distributions of member and non-member data. Critically, such distributional discrepancies are closely correlated with the extent of model overfitting: overfitting causes models to ``memorize" idiosyncrasies of their training data rather than learning generalizable patterns, thereby amplifying the distinction between the outputs of member and non-member samples.
Consequently, models afflicted by more severe overfitting will exhibit larger distributional gaps between member and non-member data, rendering them more susceptible to successful MIAs.

\subsection{Takwaway Message for practitioners}
\begin{enumerate}[leftmargin=*]
    \item Prior to employing an MIA for either adversarial exploitation or privacy auditing, a comprehensive evaluation of applicable methods is imperative. This is because the efficacy of these attacks is highly sensitive to configurations across the entire machine learning pipeline, including the chosen threat model, data distribution, model architecture, and both training and post-training algorithms.
    \item Overfitting is crucial for the performance of all MIA methods. Our benchmark indicates that the generalization gap is a consistent indicator of MIA vulnerability. We can introduce some overfitting mitigation mechanisms during training to protect data privacy.
    \item We should consider the specific scenarios when selecting MIAs. Specifically, for most cases:
    \begin{itemize}
        \item Metric MIA (Confidence and Modified Entropy version) demonstrates remarkable versatility, delivering competitive performance across various pipeline configurations under the audit mode. However, it suffers utility degradation under the attack mode, because its performance is sensitive to hyperparameter selection.
        \item Quantile MIA consistently outperforms other methods under the attack mode due to its superior performance stability under different hyper parameters.
        \item RMIA demonstrates asymmetric superiority: while it significantly outperforms other methods in low-FPR regimes (high-confidence leakage), its efficacy diminishes in high-recall screening tasks (TNR at small FNR).
        \item BlindMI One-class outperforms other MIA methods when distinguishing the membership status of the same dataset but different models.
    \end{itemize}

   % \item \textcolor{blue}{(machine unlearning take-away by cxw)} Before applying MIAs to assess machine unlearning, one should first verify whether each attack can correctly infer the known membership status of $\gD_{\mathrm{forget}}$ on the pretrained and retrained models. In our experiments, BlindMI-One-class, BlindMI-Diff-w, and MLLeaks-3 are the most consistently reliable attacks across different unlearning settings, making their MIA values better calibrated indicators for auditing privacy leakage and measuring the output-distribution distance between unlearned and retrained models.

\end{enumerate}

\section{Conclusion}

In this paper, we present a systematic evaluation of membership inference attacks (MIAs), examining their effectiveness across all stages of the machine learning lifecycle and privacy auditing pipeline. Drawing upon our extensive experimental findings, we formulate practical guidelines for selecting optimal MIAs tailored to specific operational scenarios. Ultimately, our work underscores that a rigorous, context-aware evaluation of potential MIA methodologies is indispensable before selecting an attack for practical deployment.

\section{Acknowledgments}

We thank Yuan Chen and Bofan Yang, master's students at City University of Hong Kong, for their beneficial discussion in the early stage of this work.

%% file: appendix/appendix.tex
\section{Additional Experimental Results}

\subsection{Performance of Target Model in the Experiments} \label{subsec:app_mia_scratch}
To indicate the overfitting degree of different settings, we summarize the performance of different model architectures and the performance of models trained on different datasets and training algorithms in Table \ref{tab:performance_summary}. Additionally, we provide performance summaries with fine-grained mislabel rates and training epochs in Table \ref{tab:performance_summary_mis} and \ref{tab:performance_summary_ep}, respectively.

\begin{table}[H]
    \centering
    \scriptsize
    \renewcommand{\arraystretch}{1} 
    \setlength{\tabcolsep}{3pt}     
    \caption{\small Performance summary in different architectures, datasets and training algorithms.} 
    \label{tab:performance_summary}
    % hsize 用于调整第一列相对于其他 X 列的比例
    \begin{tabularx}{\columnwidth}{>{\hsize=1.2\hsize}l | *{6}{>{\centering\arraybackslash}X}} 
    \Xhline{2\arrayrulewidth}
    & \makecell{Train\\Acc} & \makecell{Train\\Loss} & \makecell{Test\\Acc} & \makecell{Test\\Loss} & \makecell{Acc\\Gap} & \makecell{Loss\\Gap}\\
    \Xhline{2\arrayrulewidth}
    \multicolumn{7}{l}{\textit{\textbf{Different Datasets}, ResNet18, Standard}}\\
    \Xhline{1\arrayrulewidth}
    CIFAR100 & 99.98 & 0.0054 & 62.34 & 1.6833 & 37.64 & 1.6779\\
    CIFAR10 & 99.97 & 0.0020 & 89.06 & 0.4733 & 10.91 & 0.4713\\
    ImgNet100 & 80.90 & 0.7006 & 76.40 & 0.8521 & 4.50 & 0.1515\\    
    Super-C100 & 99.96 & 0.0029 & 74.58 & 1.1872 & 25.38 & 1.1843\\
    Mis-C100 & 99.96 & 0.0067 & 49.46 & 2.7693 & 50.50 & 2.7626\\
    \Xhline{1\arrayrulewidth}
    \multicolumn{7}{l}{\textit{CIFAR100, \textbf{Different Architectures}, Standard}}\\
    \Xhline{1\arrayrulewidth}
    ResNet-18 & 99.98 & 0.0054 & 62.34 & 1.6833 & 37.64 & 1.6779\\
    4-layer CNN & 64.06 & 1.2525 & 47.82 & 2.1517 & 16.24 & 0.8992\\
    VGG11 & 99.98 & 0.0079 & 61.94 & 1.5941 & 38.04 & 1.5862\\
    ResNet34 & 99.99 & 0.0031 & 60.88 & 1.7420 & 39.11 & 1.7389\\
    ResNet50 & 99.98 & 0.0029 & 61.94 & 1.8573 & 38.04 & 1.8544\\
    Swin-T & 98.85 & 0.0548 & 85.30 & 0.6903 & 13.55 & 0.6355\\
    \Xhline{1\arrayrulewidth}
    \multicolumn{7}{l}{\textit{CIFAR100, ResNet18, \textbf{Training Algorithms}}}\\
    \Xhline{1\arrayrulewidth}
    Standard & 99.98 & 0.0054 & 62.34 & 1.6833 & 37.64 & 1.6779\\
    Std(R50) & 99.98 & 0.0029 & 61.94 & 1.8573 & 38.04 & 1.8544\\
    Fine-tune & 99.99 & 0.0006 & 81.46 & 0.2462 & 18.53 & 0.2456\\
    Privacy & 95.70 & 0.2050 & 67.74 & 2.0177 & 27.96 & 1.8727\\
    \Xhline{2\arrayrulewidth}
    \end{tabularx}
\end{table}

\begin{table}[H]
    \centering
    \scriptsize
    \renewcommand{\arraystretch}{1}
    \caption{\small Performance summary in different mislabel rates.} 
    \label{tab:performance_summary_mis}
    \begin{tabularx}{\columnwidth}{c|XXXXXX}
    \Xhline{2\arrayrulewidth}
    \makecell{Mislabel\\Rate} & \makecell{Train\\Acc} & \makecell{Train\\Loss} & \makecell{Test\\Acc} & \makecell{Test\\Loss} & \makecell{Acc\\Gap} & \makecell{Loss\\Gap}\\
    \Xhline{2\arrayrulewidth}
    0 & 99.98 & 0.0054 & 62.34 & 1.6833 & 37.64 & 1.6779\\
    0.01 & 99.98 & 0.0051 & 61.88 & 1.7731 & 38.10 & 1.7680\\
    0.05 & 99.98 & 0.0056 & 54.52 & 2.3324 & 45.46 & 2.3268\\
    0.1 & 99.96 & 0.0067 & 49.46 & 2.7693 & 50.50 & 2.7626\\
    0.2 & 99.98 & 0.0076 & 38.08 & 3.5541 & 61.90 & 3.5465\\
    0.3 & 99.95 & 0.0077 & 28.56 & 4.3316 & 71.39 & 4.3239\\
    0.4 & 99.96 & 0.0072 & 20.24 & 4.8633 & 79.72 & 4.8561\\
    0.5 & 99.95 & 0.0077 & 13.96 & 5.3637 & 85.99 & 5.3560\\
    \Xhline{2\arrayrulewidth}
    \end{tabularx}
\end{table}

\begin{table}[H]
    \centering
    \scriptsize
    \renewcommand{\arraystretch}{1} % 极度压缩垂直空间
    \caption{\small Performance summary at different training epochs.} 
    \label{tab:performance_summary_ep}
    \begin{tabularx}{\columnwidth}{c|XXXXXX}
    \Xhline{2\arrayrulewidth}
    Epoch & \makecell{Train\\Acc} & \makecell{Train\\Loss} & \makecell{Test\\Acc} & \makecell{Test\\Loss} & \makecell{Acc\\Gap} & \makecell{Loss\\Gap}\\
    \Xhline{2\arrayrulewidth}
    200 & 99.98 & 0.0054 & 62.34 & 1.6833 & 37.64 & 1.6779\\
    190 & 99.98 & 0.0051 & 62.38 & 1.6935 & 37.60 & 1.6884\\
    180 & 99.97 & 0.0053 & 62.52 & 1.7020 & 37.45 & 1.6967\\
    170 & 99.96 & 0.0060 & 67.20 & 1.7033 & 37.76 & 1.6973\\
    160 & 99.96 & 0.0068 & 62.22 & 1.7070 & 37.74 & 1.7002\\
    150 & 99.96 & 0.0098 & 61.50 & 1.7345 & 38.46 & 1.7246\\
    140 & 99.92 & 0.0112 & 61.80 & 1.7590 & 38.12 & 1.7479\\
    130 & 99.84 & 0.0155 & 60.88 & 1.7741 & 38.96 & 1.7586\\
    120 & 99.62 & 0.0255 & 61.98 & 1.7527 & 37.64 & 1.7272\\
    110 & 99.02 & 0.0538 & 61.64 & 1.6987 & 37.38 & 1.6449\\
    100 & 71.39 & 0.9672 & 48.74 & 2.2259 & 22.65 & 1.2587\\
    90 & 67.45 & 1.1063 & 47.36 & 2.2759 & 20.09 & 1.1696\\
    80 & 68.92 & 1.0604 & 48.72 & 2.2110 & 20.20 & 1.1505\\
    70 & 70.27 & 0.9937 & 49.76 & 2.0981 & 20.51 & 1.1044\\
    60 & 61.20 & 1.3343 & 44.04 & 2.3934 & 17.16 & 1.0591\\
    50 & 67.82 & 1.0763 & 48.14 & 2.1618 & 19.68 & 1.0855\\
    40 & 64.04 & 1.2320 & 48.14 & 2.0953 & 15.90 & 0.8633\\
    30 & 58.26 & 1.4585 & 46.28 & 2.1908 & 11.68 & 0.7323\\
    20 & 53.98 & 1.6317 & 44.84 & 2.1480 & 9.14 & 0.5163\\
    10 & 42.63 & 2.1265 & 37.62 & 2.3711 & 5.01 & 0.2506\\
    \Xhline{2\arrayrulewidth}
    \end{tabularx}
\end{table}